\pdfoutput=1

\documentclass[11pt]{article}

\usepackage{acl}

\usepackage{times}
\usepackage{latexsym}

\usepackage[T1]{fontenc}

\usepackage[utf8]{inputenc}

\usepackage{microtype}

\usepackage{graphicx}
\usepackage{latexsym}
\usepackage{xspace}
\usepackage{caption}
\usepackage{subcaption} 
\usepackage{cprotect}
\usepackage{hhline}
\usepackage{array}
\usepackage{float}
\usepackage{xcolor}
\usepackage[group-separator={,},output-decimal-marker = {.}, group-minimum-digits={3}]{siunitx}  
\usepackage{colortbl}
\usepackage{booktabs} 
\usepackage{multirow}
\usepackage[shortlabels]{enumitem} 

\title{All You May Need for VQA are Image Captions}

\author{Soravit Changpinyo\thanks{$\quad$Equal contribution}$\quad$Doron Kukliansky\footnotemark[1]$\quad$ \\{\bf Idan Szpektor}$\quad${\bf Xi Chen}$\quad${\bf Nan Ding}$\quad${\bf Radu Soricut} \\
  Google Research \\
  \texttt{\{schangpi,doronk,szpektor,chillxichen,dingnan,rsoricut\}@google.com}}
  
\usepackage{amssymb}
\usepackage{amsmath,amsfonts}
\usepackage{amsopn}
\usepackage{bm} 
\usepackage{multirow}
\newlength\savewidth

\newcommand{\ProbOpr}[1]{\mathbb{#1}}

\newcommand{\expect}[2]{%
\ifthenelse{\equal{#2}{}}{\ProbOpr{E}_{#1}}
{\ifthenelse{\equal{#1}{}}{\ProbOpr{E}\left[#2\right]}{\ProbOpr{E}_{#1}\left[#2\right]}}} 
\newcommand{\var}[2]{%
\ifthenelse{\equal{#2}{}}{\ProbOpr{VAR}_{#1}}
{\ifthenelse{\equal{#1}{}}{\ProbOpr{VAR}\left[#2\right]}{\ProbOpr{VAR}_{#1}\left[#2\right]}}} 

\newcommand{\eat}[1]{}

\newcommand{\mypartop}[1]{\vspace{0mm}\noindent\textbf{#1}.}
\newcommand{\mypar}[1]{\vspace{0.5em}\noindent\textbf{#1}.}

\newcommand{\qsq}{$\mathrm{VQ}^{2}\!\mathrm{A}$\xspace}

\newcommand{\vqaset}{VQA2.0\xspace}
\newcommand{\gqa}{GQA\xspace}
\newcommand{\okvqa}{OKVQA\xspace}
\newcommand{\vsw}{Visual7W\xspace}
\newcommand{\cocoqa}{COCOQA\xspace}

\newcommand{\coco}{COCO\xspace}

\newcommand{\ccslong}{CC3M\xspace}

\newcommand{\qsqcc}{\qsq-CC3M\xspace}
\newcommand{\qsqcoco}{\qsq-COCO\xspace}

\newcommand{\qqex}[1]{``\emph{#1}''}
\newcommand{\qex}[1]{`\emph{#1}'}
\newcommand{\ICDS}{$D$}
\newcommand{\ICDSsize}{N}
\newcommand{\Ipair}{\mathrm{img}}
\newcommand{\Cpair}{\mathrm{cap}}
\newcommand{\AnsSize}{M}
\newcommand{\Qpair}{\mathrm{q}}
\newcommand{\Apair}{\mathrm{a}}

\newif\ifdraft
\drafttrue
\ifdraft
  \newcommand{\beer}[1]{{\color{cyan}Beer: #1}\xspace}
  \newcommand{\doron}[1]{{\color{olive}Doron: #1}\xspace}
  \newcommand{\idan}[1]{{\color{orange}Idan: #1}\xspace}
  \newcommand{\radu}[1]{{\color{red}Radu: #1}\xspace}
  \newcommand{\authore}[1]{{\color{magenta}A4 #1}\xspace}
  \newcommand{\authorf}[1]{{\color{blue}A5 #1}\xspace}
\else
  \newcommand{\beer}[1]{}
  \newcommand{\doron}[1]{}
  \newcommand{\idan}[1]{}
  \newcommand{\radu}[1]{}
  \newcommand{\authore}[1]{}
  \newcommand{\authorf}[1]{}
\fi

\begin{document}
\maketitle

\begin{abstract}
Visual Question Answering (VQA) has benefited from increasingly sophisticated models, but has not enjoyed the same level of engagement in terms of data creation. In this paper, we propose a method that automatically derives VQA examples at volume, by leveraging the abundance of existing image-caption annotations combined with neural models for textual question generation. We show that the resulting data is of high-quality. VQA models trained on our data improve state-of-the-art zero-shot accuracy by double digits and achieve a level of robustness that lacks in the same model trained on human-annotated VQA data.
\end{abstract}
\section{Introduction}
\label{sec:intro}

Visual Question Answering (VQA) is a complex multimodal task that, to be successfully modeled and evaluated, requires large amounts of annotations that are not naturally produced by existing business processes, the way translation-pair annotations \cite{guo2018effective} or image alt-text annotations \cite{cc3m} are produced.

At present, a main bottleneck for developing robust VQA systems that are useful for downstream applications, such as for visually-impaired people and in the medical and education domains,
appears to be a lack of large image-question-answer training triplets (on the order of millions).
Manual annotation of such triplets is costly, time-consuming, and prone to a variety of human biases that are difficult to account for \cite{yuan2021language}.
In addition, the brittleness of VQA systems trained on such manual annotations is well-understood and documented \cite{vqacp,tdiuc}.

\begin{figure}
    \centering
    \resizebox{0.92\linewidth}{!}{%
    \includegraphics{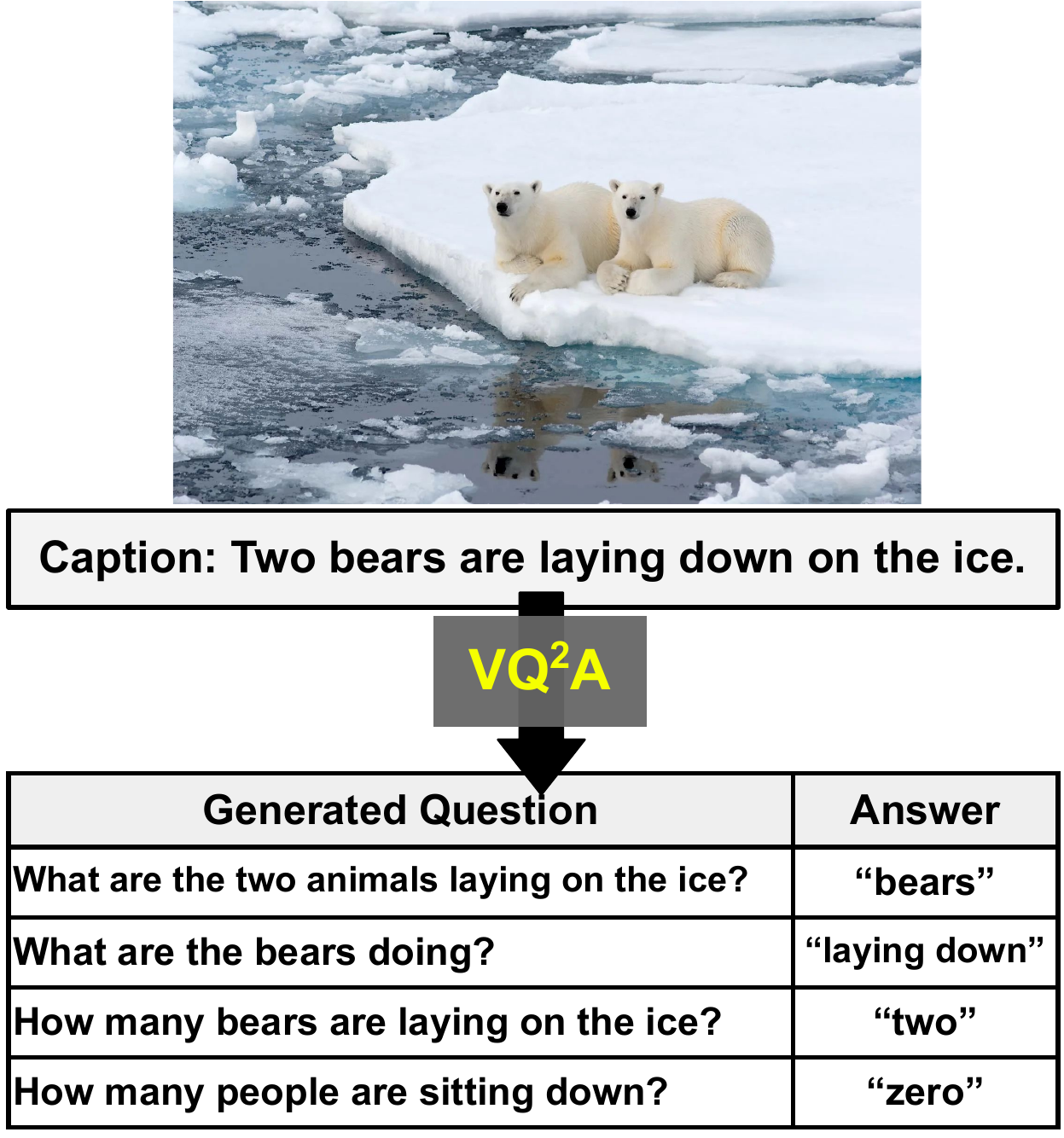}
    }
    \caption{\small Given an English caption (along with its corresponding image), our \qsq method generates high-quality question-answer pairs. These image-question-answer triplet data can be automatically produced at volume (millions of examples) and used to effectively train VQA systems.}
    \label{fig:intro}
\vspace{-15pt}
\end{figure}

To address the data limitation, we turn to a potential source for creating VQA examples: image-English caption pairs~\cite{cococap,cc3m}. Large-scale image caption datasets exist with millions \cite{cc12m}, several hundreds millions \cite{clip}, or even billions \cite{align} of examples.
Captions come mostly in the form of declarative sentences, e.g., \qqex{two bears are laying down on the ice}. Yet, the task of converting declarative captions into VQA question/answer pairs is still largely unexplored. It requires automatically inducing candidate answers fitting the VQA task, along with their respective questions based on the caption text (Fig.~\ref{fig:intro}).
We note that transforming declarative form to interrogative form plus answer(s) seems crucial, as there exists evidence that a vision-and-language model trained on declarative-language data cannot be successfully adapted or transferred ``out-of-the-box" for VQA \cite{wang2021simvlm}. 

In this paper, we explore the automatic creation of millions of VQA training data using neural models for textual question generation and question answering.
We refer to this method as \qsq, for \textbf{V}isual \textbf{Q}uestion Generation with \textbf{Q}uestion \textbf{A}nswering validation.
We demonstrate that VQA models trained on such data, with no exposure to human-annotated VQA data at all, exhibit high zero-shot performance. Our best models obtain 61.1\% accuracy on \vqaset, 52.1\% on \gqa, around 15-17 points higher than previous zero-shot state-of-the-art results, and getting close to fully-supervised performance.
In addition, taking our generated examples as a test set, we provide further evidence for the brittleness of VQA systems built with human-annotated examples, as well as evidence for the robustness of VQA systems built with the automatically-induced \qsq data.

\section{Related Work}
\label{sec:related}

\subsection{Question generation in NLP}
Question Generation (QG) is an active research topic in NLP. It is explored as a standalone task \cite{heilman2009question,nema-etal-2019-lets}, as a pre-training task for language models \cite{narayan2020qurious} and as a component in solutions for other textual tasks, such as question answering \cite{alberti-etal-2019-synthetic,puri-etal-2020-training}, information retrieval \cite{mass-etal-2020-unsupervised,gaur2021iseeq} and generation evaluation \cite{durmus-etal-2020-feqa,wang-etal-2020-asking,honovich2021q2}.
There are two main directions to QG: template-based \cite{heilman2009question,lyu2021improving,dhole2020synqg} and neural-based, with the latter achieving state-of-the-art results~\cite{alberti-etal-2019-synthetic,narayan2020qurious}.

\subsection{Question generation in computer vision}
\label{ssec:QGinVQA}

Question generation in computer vision aims at generating \emph{visual} questions about a given image (or video), either for generating questions without knowing the answer \cite{mostafazadeh-etal-2016-generating,ZhangQYYZ17,YangLLBP18,UeharaTUH18,krishna2019information}, e.g., for them to to be answered by humans, or to help improving the VQA task \cite{kafle2017data,li2018visual,shah2019cycle,Xu2021,kil2021discovering,akula2021crossvqa}, e.g., for additional evaluation and as means of data augmentation.
Such QG models are typically based on VQA triplets as training data, whose language complexity is often limited, or require the collection of visual QG data \cite{mostafazadeh-etal-2016-generating}.
We take a different approach by leveraging models trained on \emph{textual} QA datasets instead.

Multiple works leverage image captions or video transcripts as training sources \cite{ren2015exploring,banerjee2021weaqa,yang2021just,QACE}. In this approach, question-answer pairs are automatically generated from the text, ignoring the visual source, and are then combined with the related image/video to produce image-question-answer triplets.
\citet{banerjee2021weaqa} propose WeaQA, in which they generate questions from MSCOCO image captions \cite{cococap} using an improved template-based approach in \cocoqa~\cite{ren2015exploring} as well as QA-SRL methods, enhanced by paraphrasing and backtranslation for linguistic variations. \citet{QACE} similarly train a VQA model from question-answer pairs derived from MSCOCO Captions but only use noun phrases as candidate answers, focusing on using it to verify generated captions but not on the VQA task itself.
\citet{yang2021just} generate question-answer pairs from instructional video ASR transcripts, which are then coupled with the related video.

In this work, we follow this direction, investigating what requires to generate data with good coverage for the VQA task in the image domain.
We show that our neural-based textual question generation approach with captions is much more effective than previous approaches. Further, unlike previous work, we also explore automatically-curated out-of-domain image-text data sources.

\subsection{Transfer learning for and in VQA}

Existing work also explores the relationship between the image captioning task and the VQA task without question generation (Section~\ref{ssec:QGinVQA}).
\citet{fisch2020capwap} perform image captioning by anticipating visual questions (i.e., using VQA data as additional supervision and post-inference evaluation).
\citet{wu2019generating} generate question-relevant image captions to aid VQA. \citet{yang2021empirical} prompt the GPT-3~\cite{brown2020language} to answer knowledge-based visual questions based on generated captions and tags and a few VQA examples. 

Evidence suggests that image-text pre-training, especially when performed at scale, benefits vision-and-language tasks, including VQA~\cite{lu19vilbert,li19visualbert,chen20uniter,tan19lxmert,su20vlbert,lu2012in1,zhou20unified,li20oscar,zhang2021vinvl,cho2021vltt5,wang2021simvlm,yuan2021florence}. However, these approaches do not work well without fine-tuning on the downstream VQA data~\cite{wang2021simvlm}. Further, prompt-based learning and inference~\cite{liu2021pre} from a pre-trained image-text model that works for VQA is still an open research problem.
In contrast, our approach directly works with the training data, explicitly transforms them into the interrogative form of question-answer pairs.

Our focus is the \emph{zero-shot} transfer setting in WeaQA \cite{banerjee2021weaqa} in which no manually-created VQA triplets are available during training. Note that the term zero-shot here is different from the one used in \cite{teney2016zero}, in which the model still has access to manually-created VQA triplets but is evaluated with unseen questions at test time.
Similar to this, \citet{chao2018cross} explore cross-dataset VQA but they solely focus on human-annotated data along with approaches to transfer. 

\section{Textual Question Generation for VQA}
\label{sec:approach}

\begin{figure*}
    \centering
    \resizebox{0.87\linewidth}{!}{%
    \includegraphics{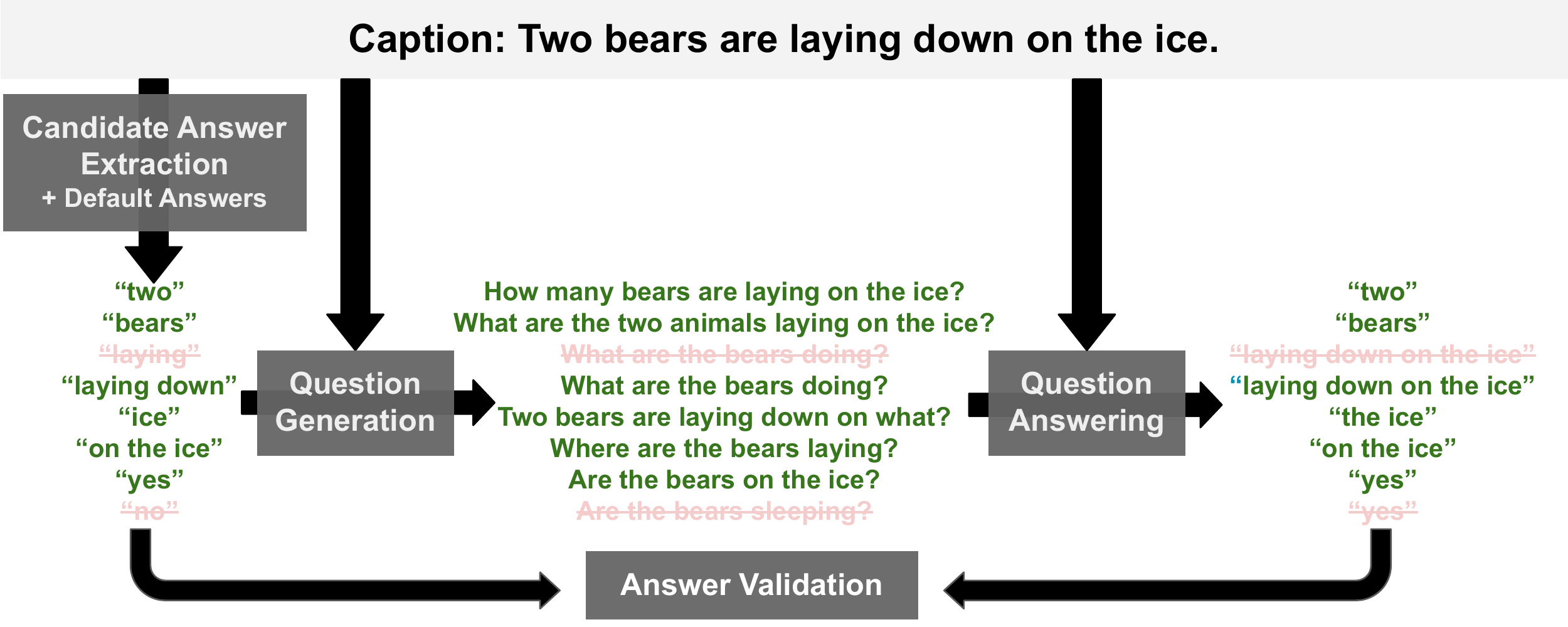}
    }
    \caption{\small \textbf{V}isual \textbf{Q}uestion Generation with \textbf{Q}uestion \textbf{A}nswering validation (\qsq) has three main stages: Candidate Answer Extraction (Section~\ref{ssec:candidate_answers}), Question Generation (Section~\ref{ssec:QG}), and Question-Answering Filtering (Question Answering + Answer Validation, Section~\ref{ssec:qa_filtering}).}
    \label{fig:approach}
\vspace{-10pt}
\end{figure*}

We study whether automatically producing VQA annotations from existing image-text resources can alleviate or completely replace the need for manual data annotation. We only focus on English in this paper.
To this end, we follow and improve upon some of the recent directions in Section~\ref{ssec:QGinVQA} on automatic question-answer generation from text.

We start with a given dataset of image-caption pairs \ICDS=$\{\Ipair_i, \Cpair_i\}_{i=1}^{\ICDSsize}$.
An important assumption we take is that the information conveyed by the caption is, in the vast majority of cases, present in the image, i.e., captions do not contain an excessive amount of external-world or personal knowledge (e.g., \qqex{\underline{my friend} at \underline{my} birthday party}). 

For each pair $\{\Ipair_i, \Cpair_i\}$, an initial set of candidate answers $\{\Apair_{i,j}\}_{j=1}^{\AnsSize_i}$ is first automatically derived from $\Cpair_i$.
For each such candidate answer, a question is generated by a neural model $\Qpair_{i,j}=QG(\Apair_{i,j}, \Cpair_i)$.
Each generated question-answer pair undergoes a validation step, and, if validated, is coupled with the corresponding image $\Ipair_i$ to induce a VQA example triplet $\{ \Ipair_i, \Qpair_{i,j}, \Apair_{i,j}\}$.

We refer to this method as \qsq (\textbf{V}isual \textbf{Q}uestion Generation with \textbf{Q}uestion \textbf{A}nswering validation). 
Figure~\ref{fig:approach} provides an overview of our approach.
We next detail the steps in \qsq.

\subsection{Candidate Answer Extraction}
\label{ssec:candidate_answers}

The only prior work on neural question generation from captions we are aware of, \citet{QACE}, focuses on noun phrases as candidate answers.
Yet, these are not enough to cover the answer types included in typical VQA benchmarks such as \vqaset (as we will show in Section~\ref{ssec:exp_vqa_zero_res}), such as boolean, attribute, and verb answers, to name a few, which are required for questions like as \qqex{Is there...}, \qqex{What color...}, \qqex{What is the dog doing}.
We present a method that covers all of these answer types.

To extract candidate answers from a given caption, we parse it using spaCy\footnote{\url{https://spacy.io/}} and then extract candidates based on the Part-of-Speech (POS) and dependency parse tree annotations, as follows:

\mypar{\textbf{Noun Phrases}} We extract all noun phrases annotated by spaCy, including named entities.

\mypar{\textbf{POS Spans}} We extract sequences that begin with an open-class POS (nouns, verbs, adjectives and adverbs), that end with an open-class POS or an adverbial particle, and that do not contain any other POS in between except closed-class POS for determiners, adpositions and conjunctions.

\mypar{\textbf{Parse Tree Spans}} We consider all sub-trees that include at least one open-class POS and no more than 3 words altogether. We only extract maximal spans, i.e., not extracting sub-trees that are fully included in other extracted sub-trees.

\mypar{\textbf{Boolean}} Boolean questions are frequent in VQA benchmarks \cite{vqa2}. Yet, \qex{yes} and \qex{no} are not found in captions, and so cannot be extracted as candidates by extracting text spans from captions. To this end, we also add \qex{yes} and \qex{no} as candidate answers and generate one question per candidate (see Section \ref{ssec:QG}).

\mypar{\textbf{How many? 0}} Captions do not normally contain mentions of \qex{zero} object counts. Hence, marking spans in a caption does not generate questions with the answer \qex{0}.
Therefore, we randomly sample a generated \qqex{How many?} question (with a non-zero answer) from a different caption and add it with the answer changed to \qex{zero} to the candidate set of the target caption.
This procedure is potentially noisy because the answer for the sampled question could be non-zero also for the target image. From a manual inspection of 200 such questions, we found this to happen infrequently -- about 4.5\%.
\ \\

\begin{table}[t]
\centering
\small
\begin{tabular}{|l|c|c|c|c|}
\hline
\multicolumn{1}{|c|}{Candidate}          &         Noun &       POS &      Parse  &     Boolean \\
\multicolumn{1}{|c|}{Answer}             &       Phrase &           &      Tree   &              \\ \hline
\qex{two}          & V            & V         &             &         \\ 
\qex{bears}        &              & V         &             &         \\ 
\qex{two bears}    & V            &           & V           &         \\ 
\qex{laying}       &              & V         &             &         \\ 
\qex{laying down}  &              & V         &             &         \\ 
\qex{ice}          &              & V         &             &         \\ 
 \qex{the ice}     & V            &           &             &         \\ 
\qex{on the ice}   &              &           & V           &         \\ 
\qex{no}           &              &           &             & V       \\ 
\qex{yes}          &              &           &             & V       \\ 
\hline
\end{tabular}
\caption{Answer candidates extracted from the sentence \qqex{two bears are laying down on the ice} and the mechanism used to extract them.}
\label{tab:qg_candidates}
\vspace{-15pt}
\end{table}

Our extraction method covers various answer candidates such as compound nouns, noun phrases, named entities, boolean answers, cardinal and ordinal numbers, verbs and their compounds, (multi-word) adjectives and prepositional phrases. Table~\ref{tab:qg_candidates} provides an example of candidate answers of various types and the mechanism used to extract them.

\begin{table*}[t]
\centering
\small
\begin{tabular}{|l|l|l|c|}
\hline
\multicolumn{1}{|c|}{Candidate Answer}          & \multicolumn{1}{|c|}{Generated Question} &       \multicolumn{1}{|c|}{Validated Answer} &      Match Score \& Result  \\
\hline
\qex{two}              & \qex{How many bears are laying on the ice?}       & \qex{two}                    & 1.0\ \ (Pass)     \\ 
\qex{bears}            & \qex{What are the two animals laying on the ice?} & \qex{bears}                  & 1.0 \ \      (Pass)     \\
\qex{two bears}        & \qex{How many bears are laying on the ice?}       & \qex{two}                    & 1.0 \ \      (Pass)     \\ 
\qex{laying}           & \qex{What are the bears doing?}                   & \qex{laying down on the ice} & 0.4 \ \      (Fail)     \\
\qex{laying down}      & \qex{What are the bears doing?}                   & \qex{laying down on the ice} & 0.7 \ \    (Pass)     \\ 
\qex{ice}              & \qex{Two bears are laying down on what?}          & \qex{the ice}                & 1.0 \ \    (Pass)     \\ 
\qex{the ice}          & \qex{Where are the bears laying?}                 & \qex{on the ice}             & 0.7 \ \    (Pass)     \\ 
\qex{on the ice}       & \qex{Where are the bears laying?}                 & \qex{on the ice}             & 1.0 \ \     (Pass)     \\ 
\qex{no}               & \qex{Are the bears sleeping?}                     & \qex{yes}                    & 0.0  \ \    (Fail)     \\ 
\qex{yes}              & \qex{Are the bears on the ice?}                   & \qex{yes}                    & 1.0  \ \    (Pass)     \\ 
\qex{zero}             & \qex{How many people are sitting down?}           & -                            &            Pass by definition     \\ 
\hline
\end{tabular}
\caption{Question/answer pairs generated from the sentence \qqex{two bears are laying down on the ice} and the filtering decision. For answer candidate \qex{zero}, no validation is performed .}
\label{tab:qg_example}
\vspace{-15pt}
\end{table*}

\subsection{Question Generation}
\label{ssec:QG}

Our question generation model, $\Qpair=QG(\Apair, \Cpair)$, takes as input a caption, $\Cpair$, and a candidate answer span within it, $\Apair$, and generates a question $\Qpair$, whose answer given the input caption is the input answer span. Importantly, the answer $\Apair$ does not need to appear verbatim in the caption, enabling the generation of questions for answer types like boolean and zero counts (see Section~\ref{ssec:candidate_answers}).

Given the advances in neural text generation, including models like T5 \cite{t5}, we choose to use a neural generation model as $QG$.
Concretely, we use a T5-XXL model and further fine-tune it on SQuAD1.1 \citep{squad} for question generation.
We take the top-scoring generated question for each caption-answer input.
We note that our QG model is trained on a question answering dataset that is not caption-specific, and therefore is not optimized for caption inputs.
From manual inspection of hundreds of generated questions, our QG model copes well with captions as input; see examples in Table~\ref{tab:qg_example} and Section~\ref{sec:data_analysis}.

\subsection{Question-Answer Filtering}
\label{ssec:qa_filtering}

Generative models may hallucinate, that is, generate content that is inconsistent with its input source \cite{alberti-etal-2019-synthetic,honovich2021q2}. To mitigate this, we follow \cite{alberti-etal-2019-synthetic} and apply round-trip consistency by answering the generated question on the caption text with a question answering model. If the answer does not match the answer candidate offered as input to the question generation model, the generated question is discarded.

We use the token-level F1 score \cite{wang-etal-2020-asking} to determine if the candidate answer and the QA model's answer is a match; If the score is above a threshold (manually set to 0.54, exemplified in Table~\ref{tab:qg_example}), the pair is a match. For question answering, we use a T5-XXL model and further fine-tune it on SQuAD2.0~\cite{squad2} and Natural Questions~\cite{nq}.

\subsection{Sources of Image/Caption Data}
\label{ssec:sources}

\begin{table}[t]
\small
\begin{center}
\begin{tabular}{@{}l|r|r|r|r@{}}
\multicolumn{1}{c|}{Dataset} & \multicolumn{2}{c|}{Image} & \multicolumn{2}{c}{VQA examples} \\ \cline{2-5}
& \multicolumn{1}{c|}{train} & \multicolumn{1}{c|}{dev} & \multicolumn{1}{c|}{train} & \multicolumn{1}{c}{dev} \\ 
\hline
\qsq{} COCO & 114.9K & 8.4K & 3.50M & 257.5K \\
\qsq{} CC3M & 3.32M & 15.8K & 13.29M & 61.2K \\ \hline
\cocoqa{} & 64.5K & 4.7K & 108.7K & 38.6K \\ \hline
\vqaset{} & 114.9K & 8.4K & 582K & 65.1K \\
\gqa{} & 82.4K & 0.4K & 1.08M & 12.6K \\
\okvqa{} & 9K & 5K & 8.3K & 4.7K \\ \hline
\end{tabular}
\vspace{-6pt}
\caption{\textbf{Sizes} of our generated \qsq{} data (top two rows) and VQA datasets used in our experiments.}
\vspace{-15pt}
\label{tab:datasets}
\end{center}
\end{table}

To gain insights on \qsq{} potential performance, we generate VQA triplets with \qsq{} from two sources of image captions: MSCOCO Captions (COCO-CAP) \cite{cococap} and Conceptual Captions (CC3M) \cite{cc3m}. COCO-CAP captions contains 123,287 images from the COCO dataset \cite{coco}, each with 5 \emph{gold} captions manually created by raters with careful guidelines. CC3M contains 3.32M images automatically-collected from the web, each with one associated alt-text which we treat as a \emph{silver} caption. 

These datasets are quite different. Both the amount and the domain of CC3M images are larger and its captions look more plausible for capturing a larger set of object/attribute/action annotations.
On the other hand, COCO-CAP's captions are cleaner and represent image content more adequately (see also Section~\ref{sec:data_analysis}).
Thus, using COCO-CAP would show the potential of training a VQA model using \qsq in a ``cleaner'' zero-shot setup, where captions are human-curated. 
Using CC3M would indicate the potential of training on noisy web image--alt-text pairs, where scaling up to billions of examples is possible.

To quantify the impact of our method, we focus on VQA classification for the \vqaset{}~\cite{vqa2}, \gqa{}~\cite{gqa}, and \okvqa{}~\cite{okvqa} benchmarks (see Section~\ref{ssec:exp_setup}).
We thus restrict our classifier to top 5,971 answers that are part of a unified answer vocabulary from these benchmarks (Appendix~\ref{apdx:implement_data_process}).
To this end, we remove triplets whose answers are not in the target answer vocabulary, and leave the study of using all generated triplets to future work. 
We then split our datasets into train/dev sets.
In particular, since the images in \vqaset are taken from COCO, we split the \coco dataset based on the standard \vqaset train/dev splits of *train2014 and minival2014~\cite{jiang2018pythia}\footnote{With the exception of \okvqa{} in which we split into train2014/val2014 to avoid using test images during training.}. For the CC3M dataset, we use the default CC3M train/dev splits~\cite{cc3m}.
For each unique image-question pair in the dev split, we construct an answer target of size 10, following \vqaset{}, by reducing or expanding the set of seed answers that occur for this image-question pair. 
Additional details are in Appendix~\ref{apdx:implement_data_process}.

Table~\ref{tab:datasets} depicts the size of the induced datasets, named \qsqcoco and \qsqcc, as well as the VQA datasets used in our experiments.

\begin{figure*}
\resizebox{\linewidth}{!}{%
\includegraphics{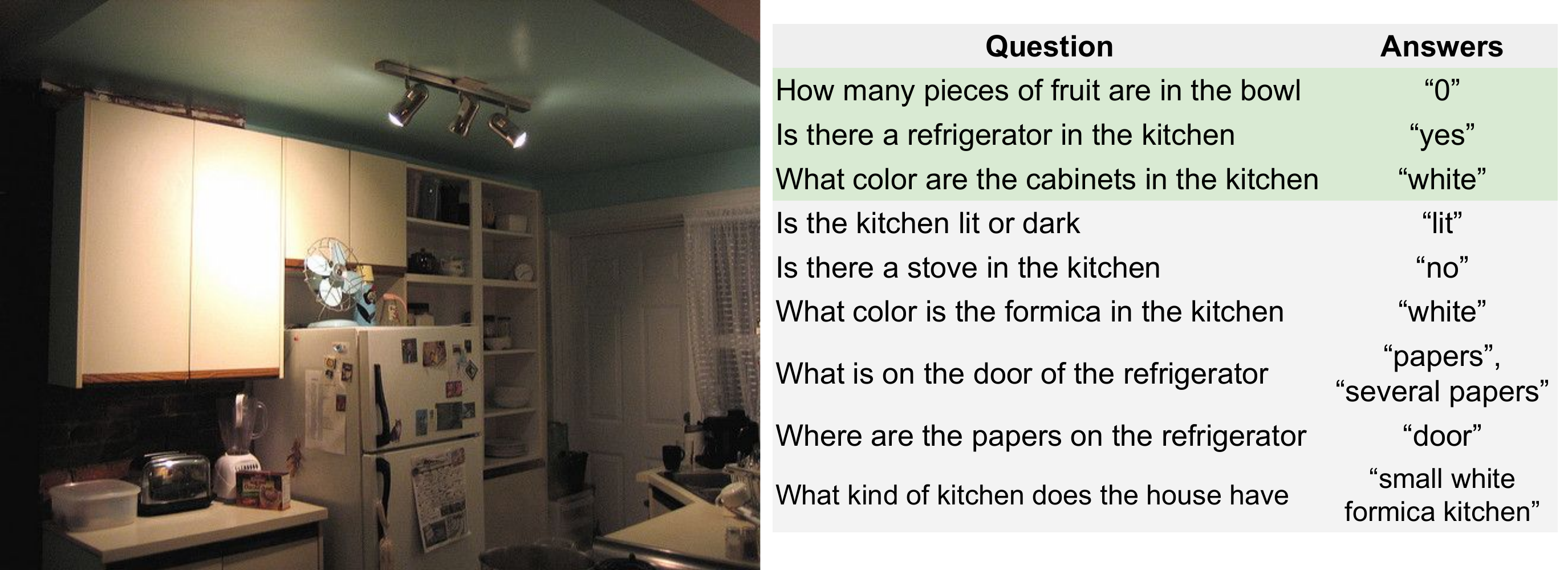}
\includegraphics{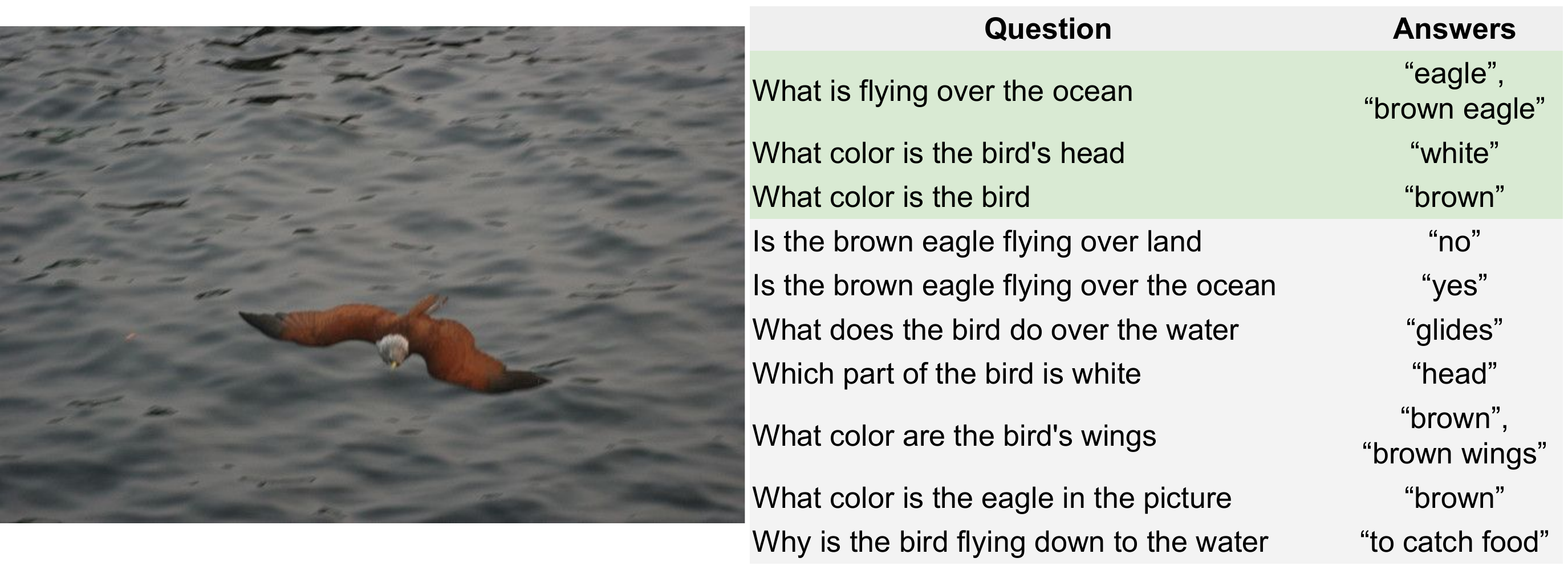}
}
\resizebox{\linewidth}{!}{%
\includegraphics{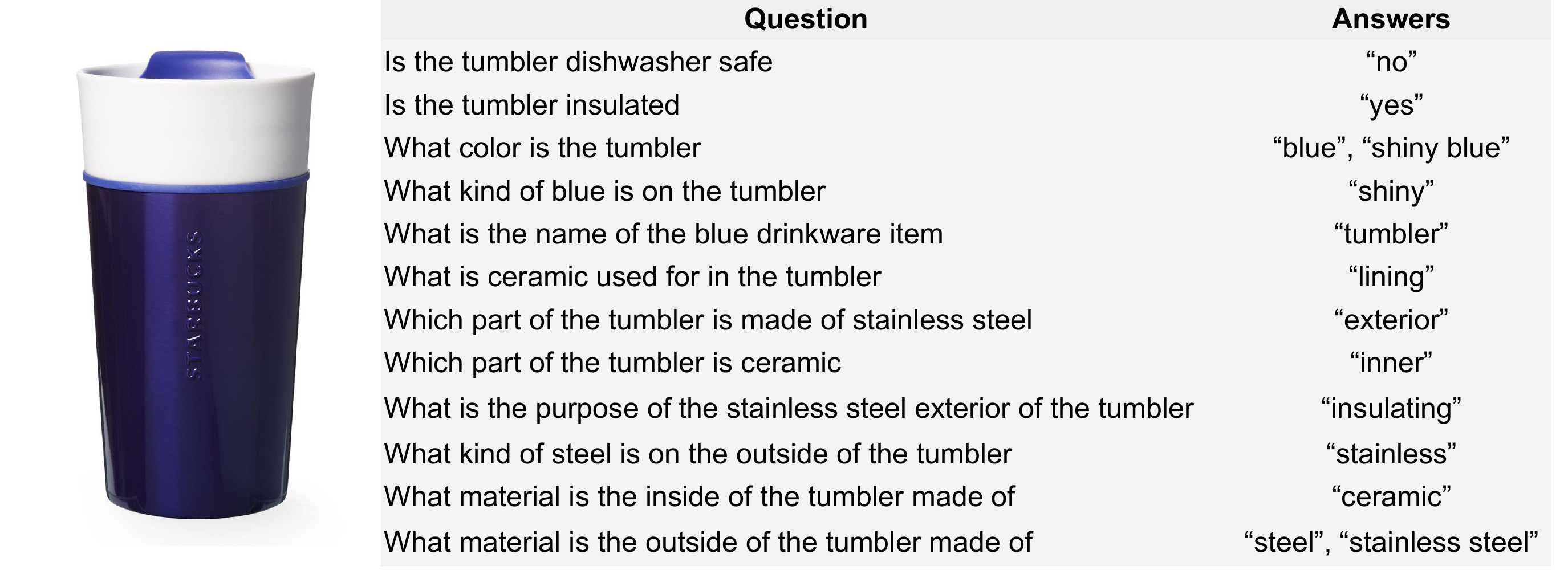}
\includegraphics{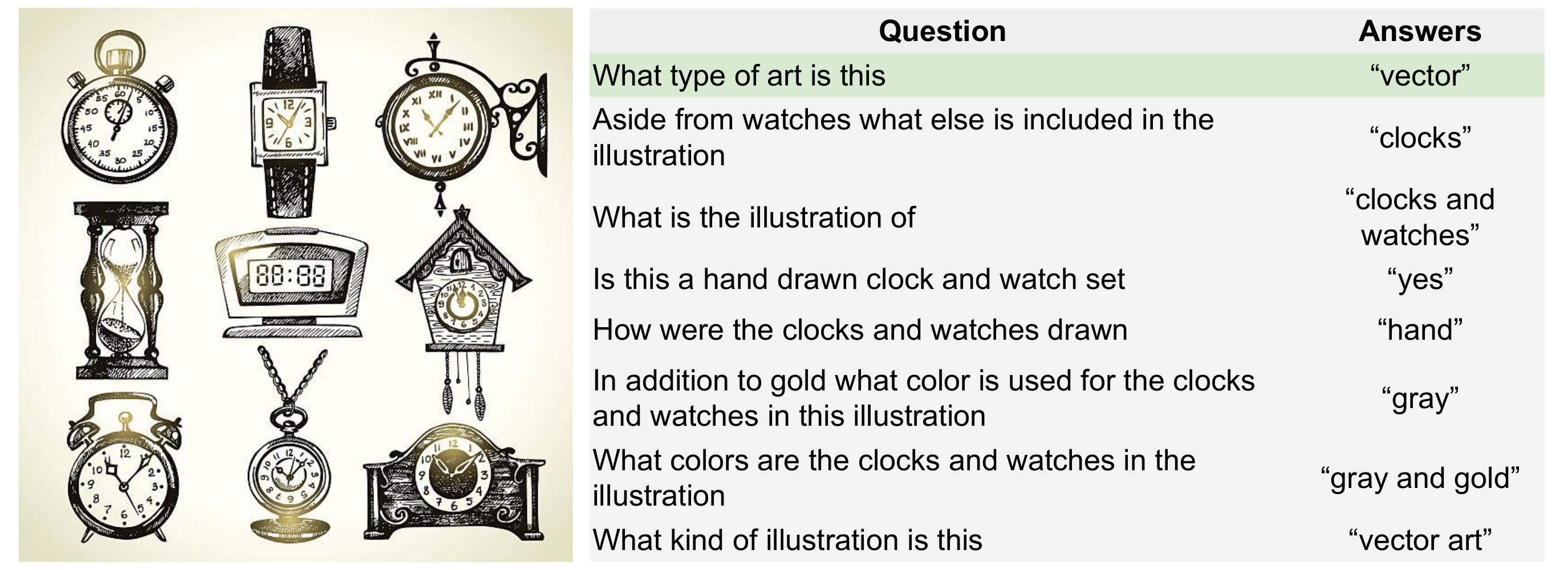}
}
\caption{\small \textbf{Examples} from \qsq{} COCO (top) and \qsq{} CC3M (bottom). Questions with the green background are also present in \vqaset{}.}
\label{fig:vq2a_qual_main}
\vspace{-15pt}
\end{figure*}

\subsection{Quality Analysis}
\label{sec:data_analysis}

To measure the quality of the generated datasets, we sampled 800 examples from each of the \qsqcoco and \qsqcc datasets.
The sample was split between four authors, who assessed whether the answer to the question in an example is justified based on the example's image.
For each dataset, 50 examples were rated by all raters, resulting in a free-margin Kappa \cite{randolph2005free} of 0.71 for \qsqcoco and 0.59 for \qsqcc, corresponding to high inter-rater agreement.
The measured percentage of valid triplets is 87.3\% for \qsqcoco and 66.0\% for \qsqcc. This shows the difference between the high-quality captions of COCO-CAP and the noisier web-based ones of \ccslong.

Fig.~\ref{fig:vq2a_qual_main} demonstrates the diversity of questions generates in the \qsq datasets. One can see that a significant amount of questions generated by \qsq for the shared \vqaset/\coco image do not appear in \vqaset.
Additional analysis and examples are in Appendix~\ref{apdx:data_analysis}.

\section{Visual Question Answering (VQA)}
\label{sec:vqa}

To assess the effectiveness of our automatic generation of VQA annotations, we perform extrinsic evaluations of the generated data by measuring its impact on a variety of established VQA benchmarks.
We first describe the model, followed by the experimental setup and the results.

\subsection{VQA Formulation and Model}
\label{sec:vqa_cls_model}

\begin{figure}
\resizebox{\linewidth}{!}{%
\includegraphics{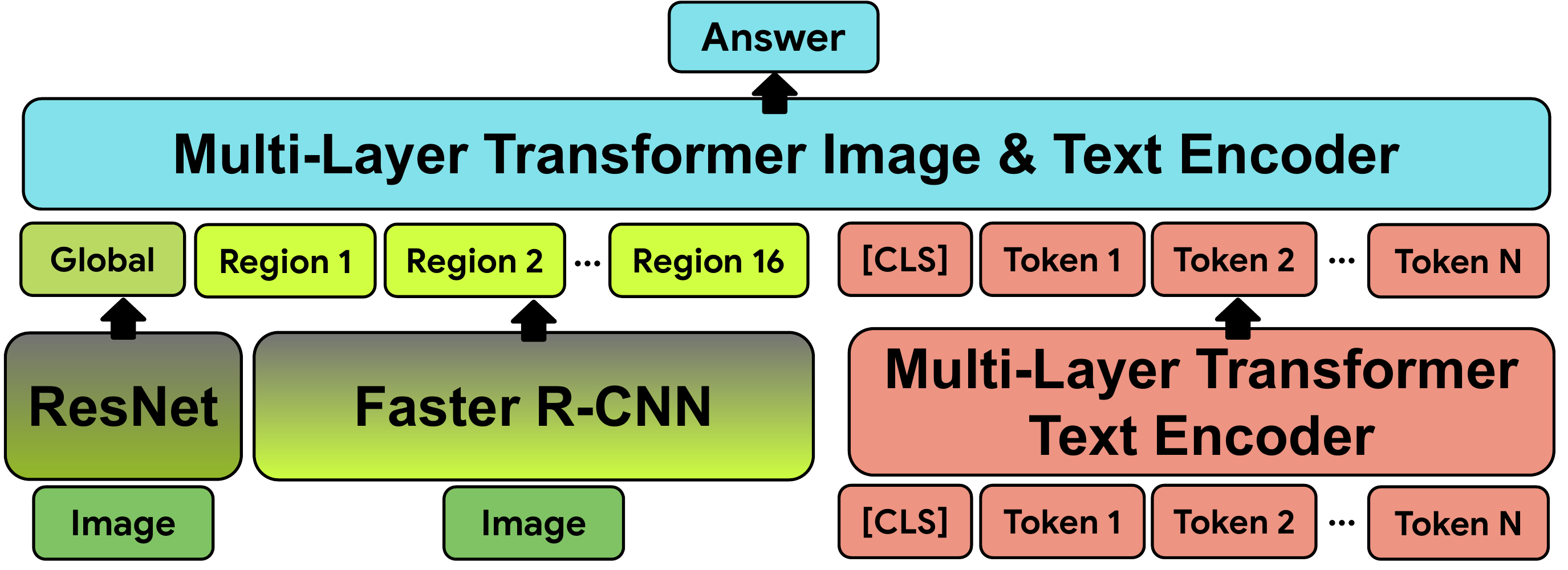}
}
\caption{\small \textbf{VQA model} used in our experiments. The text encoder is initialized from a T5-base checkpoint, while the image-text encoder is initialized from scratch. The parameters of ResNet and Faster R-CNN are frozen during VQA training.}
\label{fig:vqa_model}
\vspace{-15pt}
\end{figure}

Following the literature, we treat VQA as a classification task, i.e., vocab-based VQA. In particular, we treat our target answers as labels, where a label could be multi-token (e.g., "Christmas tree", "black and white", "play tennis").
We define our set of labels based on top answers in the training set of downstream VQA datasets, which allows for a fair comparison with most work in the VQA literature since \citet{vqa1}.

Since our work explores the impact of automatically-generated training data, we fix the VQA model architecture across all experimental conditions.
Our model fuses the input image and question (Fig.~\ref{fig:vqa_model}).
On the image side, we take global image features from ResNet-152~\cite{resnet} pre-trained on ImageNet~\cite{imagenet15} plus 16 region-of-interest image features from Faster R-CNN~\cite{fasterrcnn} pre-trained on Visual Genome~\cite{krishnavisualgenome}, following the bottom-up-features paradigm~\cite{anderson18cvpr}.
On the question side, we use the encoder of a pre-trained T5-base checkpoint~\cite{t5}.
Given the image features and the output token embeddings of the question encoder, a Transformer~\cite{transformers} fuses the multi-modal intermediate representation and classifies it into the predefined answer space.
We train the (randomly-initialized) fusing encoder and the text encoder end-to-end using standard cross-entropy loss. The parameters of both ResNet and Faster R-CNN are frozen during training.
Additional details are given in Appendix~\ref{apdx:implement_vqa}.

\subsection{Experimental Setup}
\label{ssec:exp_setup}

We consider three VQA benchmarks: \vqaset{}~\cite{vqa2}, \gqa{}~\cite{gqa}, and \okvqa{}~\cite{okvqa}.
These datasets have their own characteristics and thus test different capability of VQA models. For instance, \gqa{} puts emphasis on reasoning and \okvqa{} on external knowledge, whereas \vqaset{} is more general; \vqaset{} and \gqa{} are order-of-magnitude larger than \okvqa{}; \gqa{} is generated using a question engine while \vqaset{} and \okvqa{} are human-annotated. 

For training and evaluating on \vqaset{}, we use the standard train/dev splits *train2014 and minival2014~\cite{jiang2018pythia}. For \gqa{}, we use the balanced v1.2 and combine the train and val splits for training and use the testdev split for evaluation, following the official guideline\footnote{\url{https://cs.stanford.edu/people/dorarad/gqa/evaluate.html}} and \cite{tan19lxmert}. For \okvqa{}, we use the train/val splits for training/evaluation.
Table~\ref{tab:datasets} summarizes the sizes of the different datasets.

\mypar{Evaluation Settings and Baselines}
The main goal of our experiments is to explore the utility of our \qsq{} data for transfer learning, as training or evaluation data.

Our main focus in this paper is on zero-shot evaluation. Still, fine-tuning would provide additional insight on using our induced data for pre-training.
Therefore, following~\cite{banerjee2021weaqa}, we train VQA models on the generated \qsq{} data and then evaluate them in two settings: (i) \textbf{zero-shot} evaluation, in which we evaluate our models as-is on the dev split of \vqaset{}, \gqa{}, or \okvqa{}; and (ii) \textbf{fully-supervised} fine-tuning evaluation, in which we further fine-tune our models on the training split of \qsq, \gqa{}, or \okvqa{} before evaluating them. When training on \qsq{} data, we explore training on \qsqcoco only, \qsqcc only, and a two-stage training \qsqcc followed by \qsqcoco (\qsq{} CC3M $\xrightarrow{}$ COCO).

Our baselines, which do not use \qsq data, include (i) our VQA model trained on template-based question generation data \cocoqa{}\footnote{Train/dev based on the standard \vqaset{} train/dev splits.}~\cite{ren2015exploring}, (ii) state-of-the-art zero-shot WeaQA \cite{banerjee2021weaqa} and its fully-supervised variants, and (iii) our VQA model trained supervisely on each of the target benchmarks' training data.

\mypar{Metrics}
To be compatible with prior work, on \vqaset{} and \okvqa{} we measure the standard \emph{VQA Accuracy}. It is the average score over 9 subsets of the ground-truth 10 answers\footnote{5 targets in \okvqa{}, replicated twice~\cite{okvqa}.}, where each score is:
$min(\frac{\# answer\ occurrences}{3}, 1)$. On \gqa{}, we measure \textit{Top-1 Accuracy} against the single ground-truth answer.

\section{Results}
\label{sec:exp_vqa_res}

We report several sets of experimental results that shed light both on the accuracy and on the robustness of VQA models trained on \qsq data in this section, with additional results, analysis and ablation studies in Appendix~\ref{apdx:add_results}.

\subsection{Zero-Shot Setting}
\label{ssec:exp_vqa_zero_res}

\begin{table}[t]
\small
\begin{center}
\begin{tabular}{@{}l|c|c|c@{}}
 & \multicolumn{3}{c}{Evaluation Benchmark} \\  \cline{2-4}
\multicolumn{1}{c|}{Approach} & \vqaset{} & \gqa{} & \okvqa{} \\ \hline \hline
\multicolumn{4}{c}{Zero-shot} \\ \hline
\qsq{} COCO, nouns only & 10.5 & - & - \\
\cocoqa{} & 11.7 & 4.4 & 6.3 \\ \hline
\emph{WeaQA ZSL} & 46.8 & 33.7 & -\\ \hline
\qsq{} COCO & 60.0 & 51.3 & 18.0 \\
\qsq{} CC3M & 56.5 & 49.9 & 19.1\\
\qsq{} CC3M $\xrightarrow{}$ COCO & 61.1 & 52.1 & 19.7 \\ 
\hline
\qsq{} CC3M +D & 57.9 & 50.0 & 19.8 \\
\hline \hline
\multicolumn{4}{c}{Fully-supervised} \\ \hline
\emph{WeaQA FSL} & 65.3 & 55.2 & -\\ \hline
w/o \qsq{} data & 68.8 & 61.8 & 22.1 \\ 
\hline
w. \qsq{} COCO & 71.6 & 63.3 & 36.0 \\
w. \qsq{} CC3M & 71.3 & 63.4 & 39.0 \\
w. \qsq{} CC3M $\xrightarrow{}$ COCO & 71.4 & 64.0 & 39.3 \\ 
\hline \hline
\emph{Human performance} & 82.4$^\dagger$ & 89.3$^\ddagger$ & 82.8$^\dagger$ \\ 
\hline
\end{tabular}
\end{center}
\vspace{-3pt}
\footnotesize{
$^\dagger$ from the inter-annotator agreement of ground-truth answers.
$^\ddagger$ from \cite{gqa}.
}
\vspace{-6pt}
\caption{\textbf{\qsq{} as training data.} Accuracy in zero-shot and fully-supervised settings. All results use our architecture, except \emph{WeaQA ZSL} and \emph{WeaQA FSL}, which are the zero-shot (ZSL + Patches + Encoder) and fully-supervised (FSL + Patches + Encoder) models in \cite{banerjee2021weaqa}, respectively. +D stands for recovered raw CC3M alt-texts with digits.}
\vspace{-15pt}
\label{tab:train_vqa}
\end{table}

Table~\ref{tab:train_vqa} summarizes the outcomes of our VQA experiments on various benchmarks.
Our main result is that the \qsq{} models achieve new state-of-the-art results in the zero-shot transfer learning setting. The improvement in performance is large: to the best of our knowledge, previous state-of-the-art zero-shot accuracy was 46.8\% on \vqaset and 33.7\% on \gqa{} by WeaQA \cite{banerjee2021weaqa}, which also induces their training VQA data from COCO Captions.
Our \qsqcoco model reaches 60.0\% on \vqaset and 51.3\% on \gqa{},
an absolute improvement of +13.2\% and +17.6\%, respectively.
Even higher accuracy for the zero-shot setting -- 61.1\% (\vqaset) and 52.1\% (\gqa) -- is reached with the \qsq{} CC3M $\xrightarrow{}$ COCO model (trained first on the CC3M-derived data and then fine-tuned on the COCO-derived data), establishing new state-of-the-art results.

Training the same model architecture on the manually-constructed \vqaset and \gqa training sets in a fully-supervised manner achieves 68.8\% and 61.8\% accuracy, respectively. Hence, our results significantly close the performance gap between automatically-generated and manually-constructed training sources, indicating that the \qsq{} method may reduce the need for human curated VQA training examples.

The captions for \coco images are carefully annotated to be of high-quality \cite{cococap}.
Additionally, the \vqaset images are taken from \coco. 
To test the robustness of \qsq, we also evaluate a \qsqcc model.
While \ccslong contains more image--alt-text pairs than \coco (see Table~\ref{tab:datasets}), the images are from a different distribution and the text annotations are noisier and may represent a larger spectrum of discourse intents \cite{alikhani-etal-2020-cross}.
In spite of these differences, the gap between \coco-based and \ccslong-based \qsq{} models is not large, 60.0\% vs 56.5\% on \vqaset and 51.3\% vs. 49.9\% on \gqa.
This result strengthens our previous observation, in that it does not seem to be crucial that the starting captions are manually-annotated; it appears that ``silver'' annotations such as the ones provided by \ccslong are competitive in zero-shot VQA performance.  

To cover the types of answers present in VQA benchmarks, there is a need for thorough extraction of various answer/question types (Section~\ref{sec:approach}).
The QACE model \cite{QACE}, for example, focuses only on noun-phrases as answer types.
By analyzing the VQA2 devset, we find that only 32\% of its answers are nouns.
As such, it makes sense that, when limiting to only this answer type, the \emph{VQA Accuracy} of \qsqcoco is 10.5\%, compared to the 60\% achieved with a full coverage.
As another example, our model trained \cocoqa{} \cite{ren2015exploring}, which focuses on a few answer types and one-word answers, barely surpasses the accuracy of our \emph{COCO, nouns only} baseline.
For similar reasons, we want to be able to generate \qex{how many} questions from the \ccslong data, even though the published annotations have been stripped of digits and numerals.
To solve this problem, we recover the original captions from the \ccslong URLs,
generate questions of the type \qex{how many}, and train an additional \qsqcc+D model.
The results in Table~\ref{tab:train_vqa} show a small but consistent improvement over vanilla \qsqcc, further closing the gap between \qsq{} models using curated ``gold'' captions and noisier ``silver'' captions.

To gain further insights, we provide a breakdown of \emph{VQA Accuracy} per \vqaset{} question types in Table~\ref{tab:prefix_accuracies}.
Boolean questions are the easiest and all models perform well on them.
More challenging question types are \qex{How many?} and \qex{What is}. One reason could be  the validity of various answers, like \qqex{several} for counts. \qex{What time?} is the most difficult, probably due to lack of such information in captions.

Finally, we provide zero-shot results on the more difficult \okvqa{} benchmark.
In this setting, a supervised model reaches 22.1\% accuracy, while \qsq models in zero-shot setting achieve close to that -- 18.0\% with \coco and 19.1\% with \ccslong, while their combination reaches 19.7\%, -2.4\% shy of the supervised level.
This result also supports the conclusion that creating training data with the \qsq
method is a good replacement for small-scale supervised training data.

\begin{table}[t]
\centering
\small		
{\setlength{\tabcolsep}{0.15cm}
\begin{tabular}{@{}l|c|c|c@{}}
\hline
\multicolumn{1}{@{}c|}{Question} & \vqaset  & \qsqcoco &  \qsqcc    \\
\multicolumn{1}{@{}c|}{Prefix}   & Supervised & Zero-shot &  Zero-shot  \\ 
\hline
Boolean        &          96.3 &          93.2 &          94.2 \\
\qex{What color}      &          69.2 &          64.8 &          56.8 \\
\qex{What kind/is}       &          52.6 &          36.9 &          32.1 \\
\qex{How many}        &          49.3 &          29.4 &          19.5 \\
\qex{Where are/is}       &          38.0 &          30.0 &          25.3 \\
\qex{What does}       &          33.0 &          24.1 &          20.3 \\
\qex{What time}       &          23.6 &          11.9 &          12.7 \\
\hline
\end{tabular}}
\caption{Aggregated average Accuracy on \vqaset for the most common question types.}
\label{tab:prefix_accuracies}
\vspace{-15pt}
\end{table}

\subsection{Fully-Supervised Setting}
Another aspect of the \qsq{} method that we want to evaluate is whether it produces 
training data that is similar with the human-annotated data, or it complements it.
To this end, we perform experiments in which we first train a model using the \qsq data, and then fine-tune it in a supervised manner using the human-annotated training data. 

The results, in the \emph{Fully-supervised} part of Table~\ref{tab:train_vqa}, tell two stories.
For \vqaset and \gqa, there is a small yet consistent improvement of the fine-tuned models on top of a model trained directly on the supervised data in each benchmark (labeled \emph{w/o \qsq}).
This indicates that, at least for these two benchmarks, there is a high overlap in the nature of the signal between the human-annotated data and the \qsq{} data.

The results on \okvqa show a different trend. Here, training first with \qsq boosts performance by +17.2\% compared to supervised training without \qsq (22.1\% $\xrightarrow{}$ 39.3\%).
The small scale of the \okvqa training set (Table~\ref{tab:datasets}) certainly contributes to this effect, but it also points to another aspect: question-answer pairs that subsume world knowledge can only be made available  at-scale to models by means that are not bottlenecked by human-annotation processes.

\subsection{Robustness of Existing VQA Training Sets}

So far we have assessed the capability of models trained on \qsq{} data.
As a complementary study, we use 500 manually-validated random samples (see Section~\ref{sec:data_analysis}) from the dev part of each \qsq dataset to assess VQA robustness for various training setups.
We use the \emph{VQA Accuracy} metric for the \qsq{} datasets (10 target answers, see Section~\ref{ssec:sources}), and \emph{Top-1 Accuracy} on \cocoqa{} (one target answer).

Table~\ref{tab:eval_vqa} shows the results. The fully-supervised models (diagonal, similar training and test distributions) achieve in-domain Accuracy around 70\%, with \qsq CC3M achieving slightly higher 76.4\% Accuracy.
When tested on out-of-domain (non-diagonal), however, each model poses performance degradation at different degrees.
First, the model based on template-generated \cocoqa{} does not generalize at all.
Second, the \vqaset{} model sees significant accuracy drops, even on the \coco (44.4\%) and \cocoqa (35.9\%), which share a similar image domain with \vqaset. This result provides another evidence that progress made on the \vqaset benchmark may not reflect progress on the VQA task in full \cite{chao2018being,lebras2020adversarial}.

In contrast, both \qsq{} COCO and \qsq{} CC3M perform robustly with more modest performance drops. For instance, on \cocoqa{}, \qsq{} CC3M achieves even better performance than \vqaset{} (42.1\% vs. 35.9\%) despite being tested on out-of-domain images.
This suggests that the \qsq{} training data possesses a higher degree of question variations, provides better answer coverage, and exhibits less biases than the manually-curated \vqaset training data, at least enough to address these different benchmarks.

\begin{table}[t]
\small
\begin{center}
\begin{tabular}{@{}l|c|c|c|c@{}}
 & \multicolumn{4}{c}{Evaluation Benchmark (Acc \%)} \\  \cline{2-5}
\multicolumn{1}{c|}{Training} & COCO- & \vqaset{} & \qsq{}& \qsq{} \\ 
\multicolumn{1}{c|}{data} & QA & & COCO & CC3M \\ \hline
\cocoqa{} & {\color{gray} 70.3} & 11.7 & 13.2 & 5.8 \\ \hline
\vqaset & 35.9 & {\color{gray} 68.8} & 44.4 & 41.6 \\ \hline
\qsq{} COCO & \textbf{55.9} & \textbf{60.0} & {\color{gray} 72.6} & \textbf{56.8} \\
\qsq{} CC3M & 42.1 & 56.5 & \textbf{65.6} & {\color{gray} 76.4} \\ \hline
\end{tabular}
\end{center}
\vspace{-6pt}
\caption{\textbf{Manually-validated \qsq{} data for robustness evaluation}: Accuracy of training on "row" and tested on "column"; diagonal (gray) numbers denote supervised setting, non-diagonal numbers denote zero-shot cross-dataset setting. Best zero-shot is in \textbf{bold}.}
\vspace{-15pt}
\label{tab:eval_vqa}
\end{table}

\section{Considerations and Limitations}
\label{sec:discuss}

Automatic data generation is prone to erroneous outputs. In \qsq these may include hallucinations of the generative model, incorrect negative sampling, and bad answer span extraction. In addition, the image captions may contain details not in the image, e.g. additional details only aware to the photo taker or personal opinions, or information that is inconsistent with the image due to human mistakes and biases.
We address some of these issues automatically, filtering bad generations via question answering round-trip validation. 
In addition, the classification task itself curbs the effects of such errors through the use of a fixed answer vocabulary. Yet, for automatic generation to be more robust, additional methods to narrow down mistakes or mismatches need to be developed.

The resulting VQA model incorporates and may reinforce some of the biases and stereotypes present in the data.
For instance, it may learn that answering questions such as ``What is the gender of this person?'' is a binary choice dictated by shallow cues, or that the answer to ``For whom is this room decorated?'' depends on stereotypical features present (or not) in the room depicted in the image.
Mitigation strategies for such issues go beyond the scope of this paper, but we encourage the research community to consider addressing these issues as central for the successful deployment of this technology.
\section{Conclusions}
\label{sec:conclusions}

In this paper, we show that high-quality VQA training data can be automatically induced at scale from existing image-caption datasets. Our method, \qsq, annotates candidate answers using syntactic parsing of the captions and then derives questions for them using neural models for question generation and question answering verification.
We demonstrate that VQA models trained only on such data exhibit high zero-shot performance with new state-of-the-art results on \vqaset and \gqa.
Additionally, we provide evidence for the brittleness of VQA systems built with human-annotated examples compared to the ones built with automatically-induced image-question-answer triplets using \qsq.

For future work, we plan to explore even larger automatically-curated image-text datasets, consisting of billions of examples. In addition, we want to test the applicability of \qsq to languages other than English, for which human-annotated VQA data is scarce.

{\normalsize{
\mypar{Acknowledgments}
We would like to thank
Or Honovich, Hagai Taitelbaum and Roee Aharoni for their help with question generation,
Sebastian Goodman for his help with the VQA infrastructure,
Piyush Sharma for his help with the Conceptual Captions,
Nassim Oufattole for his early exploration of question generation,
Gal Elidan, Sasha Goldshtein, and Avinatan Hassidim for their useful feedback.}}

{\small
\bibliography{main}

\begin{thebibliography}{77}
\expandafter\ifx\csname natexlab\endcsname\relax\def\natexlab#1{#1}\fi

\bibitem[{Agrawal et~al.(2018)Agrawal, Batra, Parikh, and Kembhavi}]{vqacp}
Aishwarya Agrawal, Dhruv Batra, Devi Parikh, and Aniruddha Kembhavi. 2018.
\newblock Don't just assume; look and answer: Overcoming priors for visual
  question answering.
\newblock In \emph{CVPR}.

\bibitem[{Akula et~al.(2021)Akula, Changpinyo, Gong, Sharma, Zhu, and
  Soricut}]{akula2021crossvqa}
Arjun Akula, Soravit Changpinyo, Boqing Gong, Piyush Sharma, Song-Chun Zhu, and
  Radu Soricut. 2021.
\newblock {CrossVQA}: Scalably generating benchmarks for systematically testing
  vqa generalization.
\newblock In \emph{EMNLP}.

\bibitem[{Alberti et~al.(2019)Alberti, Andor, Pitler, Devlin, and
  Collins}]{alberti-etal-2019-synthetic}
Chris Alberti, Daniel Andor, Emily Pitler, Jacob Devlin, and Michael Collins.
  2019.
\newblock Synthetic {QA} corpora generation with roundtrip consistency.
\newblock In \emph{ACL}.

\bibitem[{Alikhani et~al.(2020)Alikhani, Sharma, Li, Soricut, and
  Stone}]{alikhani-etal-2020-cross}
Malihe Alikhani, Piyush Sharma, Shengjie Li, Radu Soricut, and Matthew Stone.
  2020.
\newblock Cross-modal coherence modeling for caption generation.
\newblock In \emph{ACL}.

\bibitem[{Anderson et~al.(2018)Anderson, He, Buehler, Teney, Johnson, Gould,
  and Zhang}]{anderson18cvpr}
Peter Anderson, Xiaodong He, Chris Buehler, Damien Teney, Mark Johnson, Stephen
  Gould, and Lei Zhang. 2018.
\newblock Bottom-up and top-down attention for image captioning and visual
  question answering.
\newblock In \emph{CVPR}.

\bibitem[{Antol et~al.(2015)Antol, Agrawal, Lu, Mitchell, Batra,
  Lawrence~Zitnick, and Parikh}]{vqa1}
Stanislaw Antol, Aishwarya Agrawal, Jiasen Lu, Margaret Mitchell, Dhruv Batra,
  C.~Lawrence~Zitnick, and Devi Parikh. 2015.
\newblock {VQA}: Visual question answering.
\newblock In \emph{ICCV}.

\bibitem[{Banerjee et~al.(2021)Banerjee, Gokhale, Yang, and
  Baral}]{banerjee2021weaqa}
Pratyay Banerjee, Tejas Gokhale, Yezhou Yang, and Chitta Baral. 2021.
\newblock {WeaQA}: Weak supervision via captions for visual question answering.
\newblock In \emph{Findings of ACL-IJCNLP}.

\bibitem[{Bras et~al.(2020)Bras, Swayamdipta, Bhagavatula, Zellers, Peters,
  Sabharwal, and Choi}]{lebras2020adversarial}
Ronan~Le Bras, Swabha Swayamdipta, Chandra Bhagavatula, Rowan Zellers,
  Matthew~E. Peters, Ashish Sabharwal, and Yejin Choi. 2020.
\newblock Adversarial filters of dataset biases.
\newblock In \emph{ICML}.

\bibitem[{Brown et~al.(2020)Brown, Mann, Ryder, Subbiah, Kaplan, Dhariwal,
  Neelakantan, Shyam, Sastry, Askell, Agarwal, Herbert-Voss, Krueger, Henighan,
  Child, Ramesh, Ziegler, Wu, Winter, Hesse, Chen, Sigler, Litwin, Gray, Chess,
  Clark, Berner, McCandlish, Radford, Sutskever, and
  Amodei}]{brown2020language}
Tom~B. Brown, Benjamin Mann, Nick Ryder, Melanie Subbiah, Jared Kaplan,
  Prafulla Dhariwal, Arvind Neelakantan, Pranav Shyam, Girish Sastry, Amanda
  Askell, Sandhini Agarwal, Ariel Herbert-Voss, Gretchen Krueger, T.~J.
  Henighan, Rewon Child, Aditya Ramesh, Daniel~M. Ziegler, Jeff Wu, Clemens
  Winter, Christopher Hesse, Mark Chen, Eric Sigler, Mateusz Litwin, Scott
  Gray, Benjamin Chess, Jack Clark, Christopher Berner, Sam McCandlish, Alec
  Radford, Ilya Sutskever, and Dario Amodei. 2020.
\newblock Language models are few-shot learners.
\newblock In \emph{NeurIPS}.

\bibitem[{Changpinyo et~al.(2021)Changpinyo, Sharma, Ding, and Soricut}]{cc12m}
Soravit Changpinyo, Piyush Sharma, Nan Ding, and Radu Soricut. 2021.
\newblock {Conceptual 12M}: Pushing web-scale image-text pre-training to
  recognize long-tail visual concepts.
\newblock In \emph{CVPR}.

\bibitem[{Chao et~al.(2018{\natexlab{a}})Chao, Hu, and Sha}]{chao2018being}
Wei-Lun Chao, Hexiang Hu, and Fei Sha. 2018{\natexlab{a}}.
\newblock Being negative but constructively: Lessons learnt from creating
  better visual question answering datasets.
\newblock In \emph{NAACL}.

\bibitem[{Chao et~al.(2018{\natexlab{b}})Chao, Hu, and Sha}]{chao2018cross}
Wei-Lun Chao, Hexiang Hu, and Fei Sha. 2018{\natexlab{b}}.
\newblock Cross-dataset adaptation for visual question answering.
\newblock In \emph{CVPR}.

\bibitem[{Chen et~al.(2015)Chen, Fang, Lin, Vedantam, Gupta, Doll{\'a}r, and
  Zitnick}]{cococap}
Xinlei Chen, Hao Fang, Tsung-Yi Lin, Ramakrishna Vedantam, Saurabh Gupta, Piotr
  Doll{\'a}r, and C.~Lawrence Zitnick. 2015.
\newblock {Microsoft COCO Captions}: Data collection and evaluation server.
\newblock \emph{arXiv preprint arXiv:1504.00325}.

\bibitem[{Chen et~al.(2020)Chen, Li, Yu, Kholy, Ahmed, Gan, Cheng, and
  Liu}]{chen20uniter}
Yen-Chun Chen, Linjie Li, Licheng Yu, Ahmed~El Kholy, Faisal Ahmed, Zhe Gan,
  Yu~Cheng, and Jingjing Liu. 2020.
\newblock {UNITER}: {Learning UNiversal Image-TExt Representations}.
\newblock In \emph{ECCV}.

\bibitem[{Cho et~al.(2021)Cho, Lei, Tan, and Bansal}]{cho2021vltt5}
Jaemin Cho, Jie Lei, Hao Tan, and Mohit Bansal. 2021.
\newblock Unifying vision-and-language tasks via text generation.
\newblock In \emph{ICML}.

\bibitem[{Dhole and Manning(2020)}]{dhole2020synqg}
Kaustubh~D. Dhole and Christopher~D. Manning. 2020.
\newblock {Syn-QG}: Syntactic and shallow semantic rules for question
  generation.
\newblock In \emph{ACL}.

\bibitem[{Durmus et~al.(2020)Durmus, He, and Diab}]{durmus-etal-2020-feqa}
Esin Durmus, He~He, and Mona Diab. 2020.
\newblock {FEQA}: A question answering evaluation framework for faithfulness
  assessment in abstractive summarization.
\newblock In \emph{ACL}.

\bibitem[{Fisch et~al.(2020)Fisch, Lee, Chang, Clark, and
  Barzilay}]{fisch2020capwap}
Adam Fisch, Kenton Lee, Ming-Wei Chang, Jonathan~H. Clark, and Regina Barzilay.
  2020.
\newblock {CapWap}: Captioning with a purpose.
\newblock In \emph{EMNLP}.

\bibitem[{Gaur et~al.(2021)Gaur, Gunaratna, Srinivasan, and
  Jin}]{gaur2021iseeq}
Manas Gaur, Kalpa Gunaratna, Vijay Srinivasan, and Hongxia Jin. 2021.
\newblock Iseeq: Information seeking question generation using dynamic
  meta-information retrieval and knowledge graphs.
\newblock \emph{arXiv preprint arXiv:2112.07622}.

\bibitem[{Goyal et~al.(2017)Goyal, Khot, Summers-Stay, Batra, and
  Parikh}]{vqa2}
Yash Goyal, Tejas Khot, Douglas Summers-Stay, Dhruv Batra, and Devi Parikh.
  2017.
\newblock Making the {V} in {VQA} matter: Elevating the role of image
  understanding in visual question answering.
\newblock In \emph{CVPR}.

\bibitem[{Guo et~al.(2018)Guo, Shen, Yang, Ge, Cer, Abrego, Stevens, Constant,
  Sung, Strope, and Kurzweil}]{guo2018effective}
Mandy Guo, Qinlan Shen, Yinfei Yang, Heming Ge, Daniel Cer, Gustavo Abrego,
  Keith Stevens, Noah Constant, Yun-Hsuan Sung, Brian Strope, and Ray Kurzweil.
  2018.
\newblock Effective parallel corpus mining using bilingual sentence embeddings.
\newblock In \emph{WMT}.

\bibitem[{He et~al.(2016)He, Zhang, Ren, and Sun}]{resnet}
Kaiming He, Xiangyu Zhang, Shaoqing Ren, and Jian Sun. 2016.
\newblock Deep residual learning for image recognition.
\newblock In \emph{CVPR}.

\bibitem[{Heilman and Smith(2009)}]{heilman2009question}
Michael Heilman and Noah~A. Smith. 2009.
\newblock Question generation via overgenerating transformations and ranking.
\newblock Technical report, Carnegie Mellon University.

\bibitem[{Honovich et~al.(2021)Honovich, Choshen, Aharoni, Neeman, Szpektor,
  and Abend}]{honovich2021q2}
Or~Honovich, Leshem Choshen, Roee Aharoni, Ella Neeman, Idan Szpektor, and Omri
  Abend. 2021.
\newblock {$Q^2$}: Evaluating factual consistency in knowledge-grounded
  dialogues via question generation and question answering.
\newblock In \emph{EMNLP}.

\bibitem[{Hudson and Manning(2019)}]{gqa}
Drew~A. Hudson and Christopher~D. Manning. 2019.
\newblock {GQA}: A new dataset for real-world visual reasoning and
  compositional question answering.
\newblock In \emph{CVPR}.

\bibitem[{Jia et~al.(2021)Jia, Yang, Xia, Chen, Parekh, Pham, Le, Sung, Li, and
  Duerig}]{align}
Chao Jia, Yinfei Yang, Ye~Xia, Yi-Ting Chen, Zarana Parekh, Hieu Pham, Quoc~V.
  Le, Yunhsuan Sung, Zhen Li, and Tom Duerig. 2021.
\newblock Scaling up visual and vision-language representation learning with
  noisy text supervision.
\newblock In \emph{ICML}.

\bibitem[{Jiang et~al.(2018)Jiang, Natarajan, Chen, Rohrbach, Batra, and
  Parikh}]{jiang2018pythia}
Yu~Jiang, Vivek Natarajan, Xinlei Chen, Marcus Rohrbach, Dhruv Batra, and Devi
  Parikh. 2018.
\newblock Pythia v0.1: the winning entry to the vqa challenge 2018.
\newblock \emph{arXiv preprint arXiv:1807.09956}.

\bibitem[{Kafle and Kanan(2017)}]{tdiuc}
Kushal Kafle and Christopher Kanan. 2017.
\newblock An analysis of visual question answering algorithms.
\newblock In \emph{ICCV}.

\bibitem[{Kafle et~al.(2017)Kafle, Yousefhussien, and Kanan}]{kafle2017data}
Kushal Kafle, Mohammed Yousefhussien, and Christopher Kanan. 2017.
\newblock Data augmentation for visual question answering.
\newblock In \emph{INLG}.

\bibitem[{Kil et~al.(2021)Kil, Zhang, Xuan, and Chao}]{kil2021discovering}
Jihyung Kil, Cheng Zhang, Dong Xuan, and Wei-Lun Chao. 2021.
\newblock Discovering the unknown knowns: Turning implicit knowledge in the
  dataset into explicit training examples for visual question answering.
\newblock In \emph{EMNLP}.

\bibitem[{Krishna et~al.(2019)Krishna, Bernstein, and
  Fei-Fei}]{krishna2019information}
Ranjay Krishna, Michael Bernstein, and Li~Fei-Fei. 2019.
\newblock Information maximizing visual question generation.
\newblock In \emph{CVPR}.

\bibitem[{Krishna et~al.(2017)Krishna, Zhu, Groth, Johnson, Hata, Kravitz,
  Chen, Kalantidis, Li, Shamma, Bernstein, and Fei-Fei}]{krishnavisualgenome}
Ranjay Krishna, Yuke Zhu, Oliver Groth, Justin Johnson, Kenji Hata, Joshua
  Kravitz, Stephanie Chen, Yannis Kalantidis, Li-Jia Li, David~A. Shamma,
  Michael Bernstein, and Li~Fei-Fei. 2017.
\newblock {Visual Genome}: Connecting language and vision using crowdsourced
  dense image annotations.
\newblock \emph{IJCV}, 123(1):32--73.

\bibitem[{Kwiatkowski et~al.(2019)Kwiatkowski, Palomaki, Redfield, Collins,
  Parikh, Alberti, Epstein, Polosukhin, Devlin, Lee, , Toutanova, Jones,
  Kelcey, Chang, Dai, Uszkoreit, Le, and Petrov}]{nq}
Tom Kwiatkowski, Jennimaria Palomaki, Olivia Redfield, Michael Collins, Ankur
  Parikh, Chris Alberti, Danielle Epstein, Illia Polosukhin, Jacob Devlin,
  Kenton Lee, , Kristina Toutanova, Llion Jones, Matthew Kelcey, Ming-Wei
  Chang, Andrew~M. Dai, Jakob Uszkoreit, Quoc Le, and Slav Petrov. 2019.
\newblock {Natural Questions}: a benchmark for question answering research.
\newblock \emph{TACL}, 7:453--466.

\bibitem[{Lee et~al.(2021)Lee, Scialom, Yoon, Dernoncourt, and Jung}]{QACE}
Hwanhee Lee, Thomas Scialom, Seunghyun Yoon, Franck Dernoncourt, and Kyomin
  Jung. 2021.
\newblock {QACE}: Asking questions to evaluate an image caption.
\newblock In \emph{Findings of EMNLP}.

\bibitem[{Li et~al.(2019)Li, Yatskar, Yin, Hsieh, and Chang}]{li19visualbert}
Liunian~Harold Li, Mark Yatskar, Da~Yin, Cho-Jui Hsieh, and Kai-Wei Chang.
  2019.
\newblock {VisualBERT}: A simple and performant baseline for vision and
  language.
\newblock \emph{arXiv preprint arXiv:1908.03557}.

\bibitem[{Li et~al.(2020)Li, Yin, Li, Hu, Zhang, Zhang, Wang, Hu, Dong, Wei,
  Choi, and Gao}]{li20oscar}
Xiujun Li, Xi~Yin, Chunyuan Li, Xiaowei Hu, Pengchuan Zhang, Lei Zhang, Lijuan
  Wang, Houdong Hu, Li~Dong, Furu Wei, Yejin Choi, and Jianfeng Gao. 2020.
\newblock Oscar: Object-semantics aligned pre-training for vision-language
  tasks.
\newblock In \emph{ECCV}.

\bibitem[{Li et~al.(2018)Li, Duan, Zhou, Chu, Ouyang, Wang, and
  Zhou}]{li2018visual}
Yikang Li, Nan Duan, Bolei Zhou, Xiao Chu, Wanli Ouyang, Xiaogang Wang, and
  Ming Zhou. 2018.
\newblock Visual question generation as dual task of visual question answering.
\newblock In \emph{CVPR}.

\bibitem[{Lin et~al.(2014)Lin, Maire, Belongie, Bourdev, Girshick, Hays,
  Perona, Ramanan, Zitnick, and Doll\'{a}r}]{coco}
Tsung-Yi Lin, Michael Maire, Serge Belongie, Lubomir Bourdev, Ross Girshick,
  James Hays, Pietro Perona, Deva Ramanan, C.~Lawrence Zitnick, and Piotr
  Doll\'{a}r. 2014.
\newblock Microsoft {COCO}: Common objects in context.
\newblock In \emph{ECCV}.

\bibitem[{Liu et~al.(2021)Liu, Yuan, Fu, Jiang, Hayashi, and
  Neubig}]{liu2021pre}
Pengfei Liu, Weizhe Yuan, Jinlan Fu, Zhengbao Jiang, Hiroaki Hayashi, and
  Graham Neubig. 2021.
\newblock Pre-train, prompt, and predict: A systematic survey of prompting
  methods in natural language processing.
\newblock \emph{arXiv preprint arXiv:2107.13586}.

\bibitem[{Lu et~al.(2019)Lu, Batra, Parikh, and Lee}]{lu19vilbert}
Jiasen Lu, Dhruv Batra, Devi Parikh, and Stefan Lee. 2019.
\newblock {ViLBERT}: Pretraining task-agnostic visiolinguistic representations
  for vision-and-language tasks.
\newblock In \emph{NeurIPS}.

\bibitem[{Lu et~al.(2020)Lu, Goswami, Rohrbach, Parikh, and Lee}]{lu2012in1}
Jiasen Lu, Vedanuj Goswami, Marcus Rohrbach, Devi Parikh, and Stefan Lee. 2020.
\newblock 12-in-1: Multi-task vision and language representation learning.
\newblock In \emph{CVPR}.

\bibitem[{Lyu et~al.(2021)Lyu, Shang, Graham, Foster, Jiang, and
  Liu}]{lyu2021improving}
Chenyang Lyu, Lifeng Shang, Yvette Graham, Jennifer Foster, Xin Jiang, and Qun
  Liu. 2021.
\newblock Improving unsupervised question answering via summarization-informed
  question generation.
\newblock In \emph{EMNLP}.

\bibitem[{Marino et~al.(2019)Marino, Rastegari, Farhadi, and Mottaghi}]{okvqa}
Kenneth Marino, Mohammad Rastegari, Ali Farhadi, and Roozbeh Mottaghi. 2019.
\newblock {OK-VQA}: A visual question answering benchmark requiring external
  knowledge.
\newblock In \emph{CVPR}.

\bibitem[{Mass et~al.(2020)Mass, Carmeli, Roitman, and
  Konopnicki}]{mass-etal-2020-unsupervised}
Yosi Mass, Boaz Carmeli, Haggai Roitman, and David Konopnicki. 2020.
\newblock Unsupervised {FAQ} retrieval with question generation and {BERT}.
\newblock In \emph{ACL}.

\bibitem[{Mostafazadeh et~al.(2016)Mostafazadeh, Misra, Devlin, Mitchell, He,
  and Vanderwende}]{mostafazadeh-etal-2016-generating}
Nasrin Mostafazadeh, Ishan Misra, Jacob Devlin, Margaret Mitchell, Xiaodong He,
  and Lucy Vanderwende. 2016.
\newblock Generating natural questions about an image.
\newblock In \emph{ACL}.

\bibitem[{Narayan et~al.(2020)Narayan, Simoes, Ma, Craighead, and
  Mcdonald}]{narayan2020qurious}
Shashi Narayan, Gon{\c{c}}alo Simoes, Ji~Ma, Hannah Craighead, and Ryan
  Mcdonald. 2020.
\newblock {QURIOUS}: Question generation pretraining for text generation.
\newblock \emph{arXiv preprint arXiv:2004.11026}.

\bibitem[{Nema et~al.(2019)Nema, Mohankumar, Khapra, Srinivasan, and
  Ravindran}]{nema-etal-2019-lets}
Preksha Nema, Akash~Kumar Mohankumar, Mitesh~M. Khapra, Balaji~Vasan
  Srinivasan, and Balaraman Ravindran. 2019.
\newblock Let{'}s ask again: Refine network for automatic question generation.
\newblock In \emph{EMNLP-IJCNLP}.

\bibitem[{Puri et~al.(2020)Puri, Spring, Shoeybi, Patwary, and
  Catanzaro}]{puri-etal-2020-training}
Raul Puri, Ryan Spring, Mohammad Shoeybi, Mostofa Patwary, and Bryan Catanzaro.
  2020.
\newblock Training question answering models from synthetic data.
\newblock In \emph{EMNLP}.

\bibitem[{Radford et~al.(2021)Radford, Kim, Hallacy, Ramesh, Goh, Agarwal,
  Sastry, Askell, Mishkin, Clark, Krueger, and Sutskever}]{clip}
Alec Radford, Jong~Wook Kim, Chris Hallacy, Aditya Ramesh, Gabriel Goh,
  Sandhini Agarwal, Girish Sastry, Amanda Askell, Pamela Mishkin, Jack Clark,
  Gretchen Krueger, and Ilya Sutskever. 2021.
\newblock Learning transferable visual models from natural language
  supervision.
\newblock In \emph{ICML}.

\bibitem[{Raffel et~al.(2020)Raffel, Shazeer, Roberts, Lee, Narang, Matena,
  Zhou, Li, and Liu}]{t5}
Colin Raffel, Noam Shazeer, Adam Roberts, Katherine Lee, Sharan Narang, Michael
  Matena, Yanqi Zhou, Wei Li, and Peter~J. Liu. 2020.
\newblock Exploring the limits of transfer learning with a unified text-to-text
  transformer.
\newblock \emph{JMLR}.

\bibitem[{Rajpurkar et~al.(2018)Rajpurkar, Jia, and Liang}]{squad2}
Pranav Rajpurkar, Robin Jia, and Percy Liang. 2018.
\newblock Know what you don't know: Unanswerable questions for {SQuAD}.
\newblock In \emph{ACL}.

\bibitem[{Rajpurkar et~al.(2016)Rajpurkar, Zhang, Lopyrev, and Liang}]{squad}
Pranav Rajpurkar, Jian Zhang, Konstantin Lopyrev, and Percy Liang. 2016.
\newblock {SQuAD}: 100,000+ questions for machine comprehension of text.
\newblock In \emph{EMNLP}.

\bibitem[{Randolph(2005)}]{randolph2005free}
Justus~J. Randolph. 2005.
\newblock Free-marginal multirater kappa (multirater$\kappa_{\text{free}}$): An
  alternative to {Fleiss'} fixed-marginal multirater kappa.
\newblock \emph{Joensuu Learning and Instruction Symposium}.

\bibitem[{Ren et~al.(2015{\natexlab{a}})Ren, Kiros, and
  Zemel}]{ren2015exploring}
Mengye Ren, Ryan Kiros, and Richard Zemel. 2015{\natexlab{a}}.
\newblock Exploring models and data for image question answering.
\newblock In \emph{NIPS}.

\bibitem[{Ren et~al.(2015{\natexlab{b}})Ren, He, Girshick, and
  Sun}]{fasterrcnn}
Shaoqing Ren, Kaiming He, Ross Girshick, and Jian Sun. 2015{\natexlab{b}}.
\newblock {Faster R-CNN}: Towards real-time object detection with region
  proposal networks.
\newblock In \emph{NIPS}.

\bibitem[{Russakovsky et~al.(2015)Russakovsky, Deng, Su, Krause, Satheesh, Ma,
  Huang, Karpathy, Khosla, Bernstein, Berg, and Fei-Fei}]{imagenet15}
Olga Russakovsky, Jia Deng, Hao Su, Jonathan Krause, Sanjeev Satheesh, Sean Ma,
  Zhiheng Huang, Andrej Karpathy, Aditya Khosla, Michael Bernstein,
  Alexander~C. Berg, and Li~Fei-Fei. 2015.
\newblock {ImageNet} large scale visual recognition challenge.
\newblock \emph{IJCV}, 115(3):211--252.

\bibitem[{Shah et~al.(2019)Shah, Chen, Rohrbach, and Parikh}]{shah2019cycle}
Meet Shah, Xinlei Chen, Marcus Rohrbach, and Devi Parikh. 2019.
\newblock Cycle-consistency for robust visual question answering.
\newblock In \emph{CVPR}.

\bibitem[{Sharma et~al.(2018)Sharma, Ding, Goodman, and Soricut}]{cc3m}
Piyush Sharma, Nan Ding, Sebastian Goodman, and Radu Soricut. 2018.
\newblock {Conceptual Captions}: A cleaned, hypernymed, image alt-text dataset
  for automatic image captioning.
\newblock In \emph{ACL}.

\bibitem[{Shazeer and Stern(2018)}]{adafactor}
Noam Shazeer and Mitchell Stern. 2018.
\newblock Adafactor: Adaptive learning rates with sublinear memory cost.
\newblock In \emph{ICLR}.

\bibitem[{Singh et~al.(2020)Singh, Goswami, Natarajan, Jiang, Chen, Shah,
  Rohrbach, Batra, and Parikh}]{singh2020mmf}
Amanpreet Singh, Vedanuj Goswami, Vivek Natarajan, Yu~Jiang, Xinlei Chen, Meet
  Shah, Marcus Rohrbach, Dhruv Batra, and Devi Parikh. 2020.
\newblock {MMF}: A multimodal framework for vision and language research.
\newblock \url{https://github.com/facebookresearch/mmf}.

\bibitem[{Su et~al.(2020)Su, Zhu, Cao, Li, Lu, Wei, and Dai}]{su20vlbert}
Weijie Su, Xizhou Zhu, Yue Cao, Bin Li, Lewei Lu, Furu Wei, and Jifeng Dai.
  2020.
\newblock {VL-BERT}: Pre-training of generic visual-linguistic representations.
\newblock In \emph{ICLR}.

\bibitem[{Tan and Bansal(2019)}]{tan19lxmert}
Hao Tan and Mohit Bansal. 2019.
\newblock {LXMERT}: Learning cross-modality encoder representations from
  transformers.
\newblock In \emph{EMNLP-IJCNLP}.

\bibitem[{Teney and Hengel(2016)}]{teney2016zero}
Damien Teney and Anton van~den Hengel. 2016.
\newblock Zero-shot visual question answering.
\newblock \emph{arXiv preprint arXiv:1611.05546}.

\bibitem[{Uehara et~al.(2018)Uehara, Tejero{-}de{-}Pablos, Ushiku, and
  Harada}]{UeharaTUH18}
Kohei Uehara, Antonio Tejero{-}de{-}Pablos, Yoshitaka Ushiku, and Tatsuya
  Harada. 2018.
\newblock Visual question generation for class acquisition of unknown objects.
\newblock In \emph{ECCV}.

\bibitem[{Vaswani et~al.(2017)Vaswani, Shazeer, Parmar, Uszkoreit, Jones,
  Gomez, Kaiser, and Polosukhin}]{transformers}
Ashish Vaswani, Noam Shazeer, Niki Parmar, Jakob Uszkoreit, Llion Jones,
  Aidan~N. Gomez, Lukasz Kaiser, and Illia Polosukhin. 2017.
\newblock Attention is all you need.
\newblock In \emph{NeurIPS}.

\bibitem[{Wang et~al.(2020)Wang, Cho, and Lewis}]{wang-etal-2020-asking}
Alex Wang, Kyunghyun Cho, and Mike Lewis. 2020.
\newblock Asking and answering questions to evaluate the factual consistency of
  summaries.
\newblock In \emph{ACL}.

\bibitem[{Wang et~al.(2021)Wang, Yu, Yu, Dai, Tsvetkov, and
  Cao}]{wang2021simvlm}
Zirui Wang, Jiahui Yu, Adams~Wei Yu, Zihang Dai, Yulia Tsvetkov, and Yuan Cao.
  2021.
\newblock {SimVLM}: Simple visual language model pretraining with weak
  supervision.
\newblock \emph{arXiv preprint arXiv:2108.10904}.

\bibitem[{Wu et~al.(2019)Wu, Hu, and Mooney}]{wu2019generating}
Jialin Wu, Zeyuan Hu, and Raymond~J. Mooney. 2019.
\newblock Generating question relevant captions to aid visual question
  answering.
\newblock In \emph{NeurIPS}.

\bibitem[{Xu et~al.(2021)Xu, Wang, Yang, Hanjalic, and Shen}]{Xu2021}
Xing Xu, Tan Wang, Yang Yang, Alan Hanjalic, and Heng~Tao Shen. 2021.
\newblock Radial graph convolutional network for visual question generation.
\newblock \emph{IEEE Transactions on Neural Networks and Learning Systems},
  32(4):1654--1667.

\bibitem[{Yang et~al.(2021{\natexlab{a}})Yang, Miech, Sivic, Laptev, and
  Schmid}]{yang2021just}
Antoine Yang, Antoine Miech, Josef Sivic, Ivan Laptev, and Cordelia Schmid.
  2021{\natexlab{a}}.
\newblock Just ask: Learning to answer questions from millions of narrated
  videos.
\newblock In \emph{ICCV}.

\bibitem[{Yang et~al.(2018)Yang, Lu, Lee, Batra, and Parikh}]{YangLLBP18}
Jianwei Yang, Jiasen Lu, Stefan Lee, Dhruv Batra, and Devi Parikh. 2018.
\newblock Visual curiosity: Learning to ask questions to learn visual
  recognition.
\newblock In \emph{CoRL}.

\bibitem[{Yang et~al.(2021{\natexlab{b}})Yang, Gan, Wang, Hu, Lu, Liu, and
  Wang}]{yang2021empirical}
Zhengyuan Yang, Zhe Gan, Jianfeng Wang, Xiaowei Hu, Yumao Lu, Zicheng Liu, and
  Lijuan Wang. 2021{\natexlab{b}}.
\newblock {An empirical study of GPT-3 for few-shot knowledge-based VQA}.
\newblock \emph{arXiv preprint arXiv:2109.05014}.

\bibitem[{Yuan(2021)}]{yuan2021language}
Desen Yuan. 2021.
\newblock Language bias in visual question answering: A survey and taxonomy.
\newblock \emph{arXiv preprint arXiv:2111.08531}.

\bibitem[{Yuan et~al.(2021)Yuan, Chen, Chen, Codella, Dai, Gao, Hu, Huang, Li,
  Li, Liu, Liu, Liu, Lu, Shi, Wang, Wang, Xiao, Xiao, Yang, Zeng, Zhou, and
  Zhang}]{yuan2021florence}
Lu~Yuan, Dongdong Chen, Yi-Ling Chen, Noel Codella, Xiyang Dai, Jianfeng Gao,
  Houdong Hu, Xuedong Huang, Boxin Li, Chunyuan Li, Ce~Liu, Mengchen Liu,
  Zicheng Liu, Yumao Lu, Yu~Shi, Lijuan Wang, Jianfeng Wang, Bin Xiao, Zhen
  Xiao, Jianwei Yang, Michael Zeng, Luowei Zhou, and Pengchuan Zhang. 2021.
\newblock Florence: A new foundation model for computer vision.
\newblock \emph{arXiv preprint arXiv:2111.11432}.

\bibitem[{Zhang et~al.(2021)Zhang, Li, Hu, Yang, Zhang, Wang, Choi, and
  Gao}]{zhang2021vinvl}
Pengchuan Zhang, Xiujun Li, Xiaowei Hu, Jianwei Yang, Lei Zhang, Lijuan Wang,
  Yejin Choi, and Jianfeng Gao. 2021.
\newblock {VinVL}: Revisiting visual representations in vision-language models.
\newblock In \emph{CVPR}.

\bibitem[{Zhang et~al.(2017)Zhang, Qu, You, Yang, and Zhang}]{ZhangQYYZ17}
Shijie Zhang, Lizhen Qu, Shaodi You, Zhenglu Yang, and Jiawan Zhang. 2017.
\newblock Automatic generation of grounded visual questions.
\newblock In \emph{IJCAI}.

\bibitem[{Zhou et~al.(2020)Zhou, Palangi, Zhang, Hu, Corso, and
  Gao}]{zhou20unified}
Luowei Zhou, Hamid Palangi, Lei Zhang, Houdong Hu, Jason~J. Corso, and Jianfeng
  Gao. 2020.
\newblock Unified vision-language pre-training for image captioning and {VQA}.
\newblock In \emph{AAAI}.

\end{thebibliography}
\bibliographystyle{acl_natbib}
}
\newpage

\appendix

\begin{figure*}
\resizebox{\linewidth}{!}{%
\includegraphics{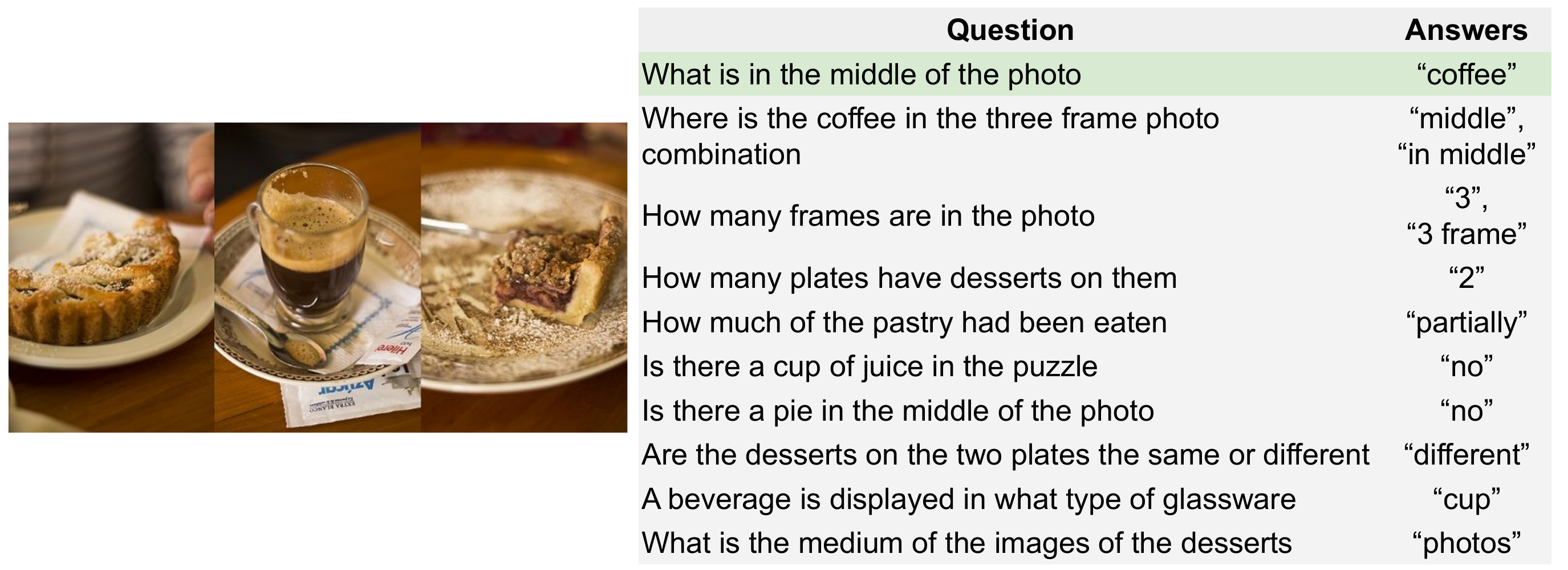}
\includegraphics{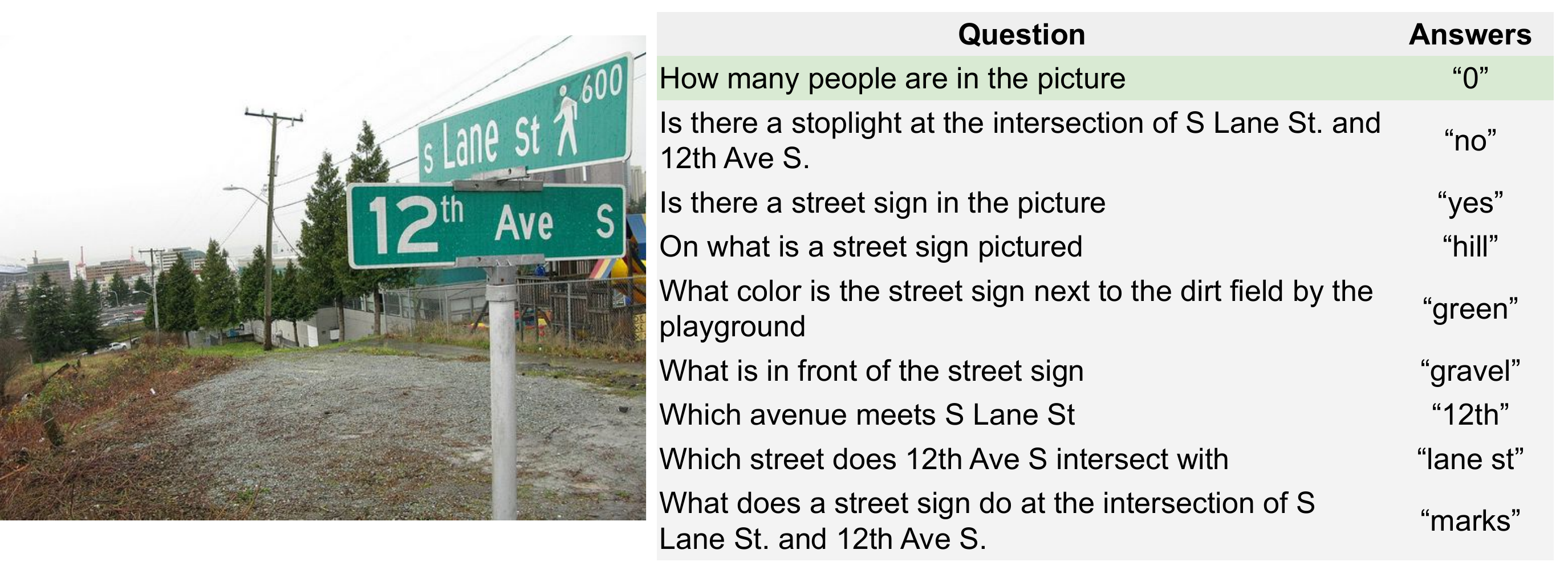}
}
\resizebox{\linewidth}{!}{%
\includegraphics{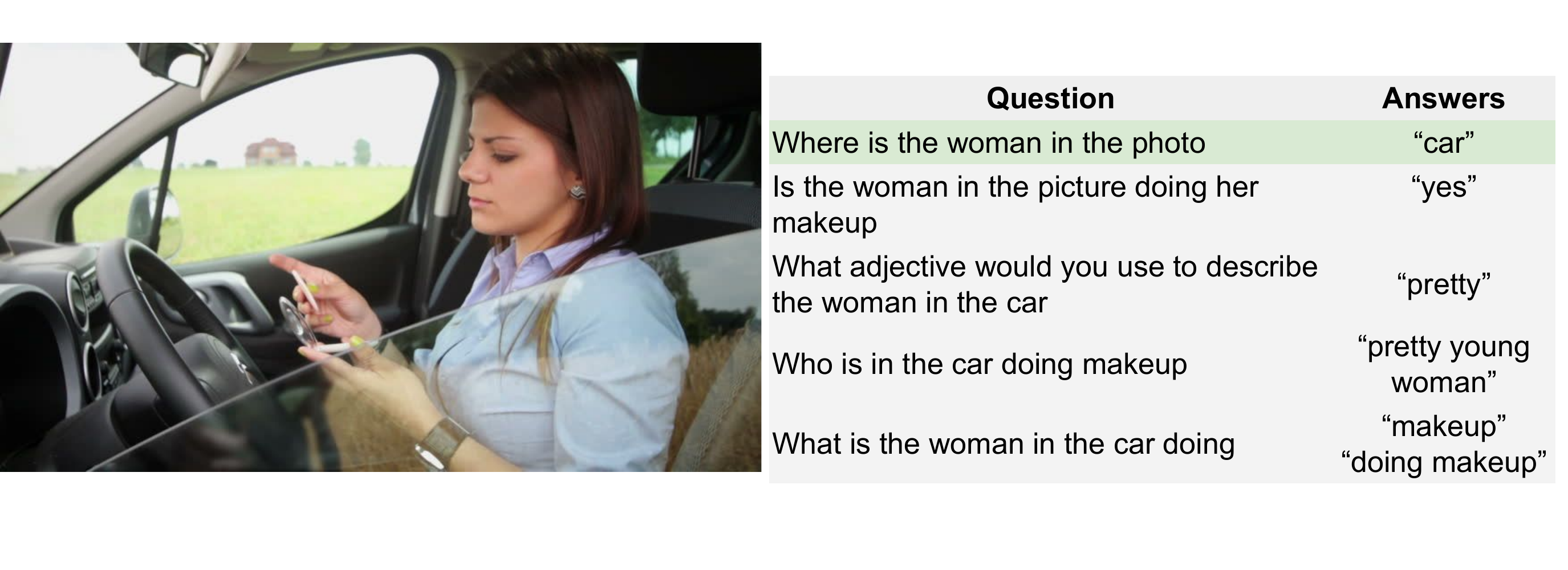}
\includegraphics{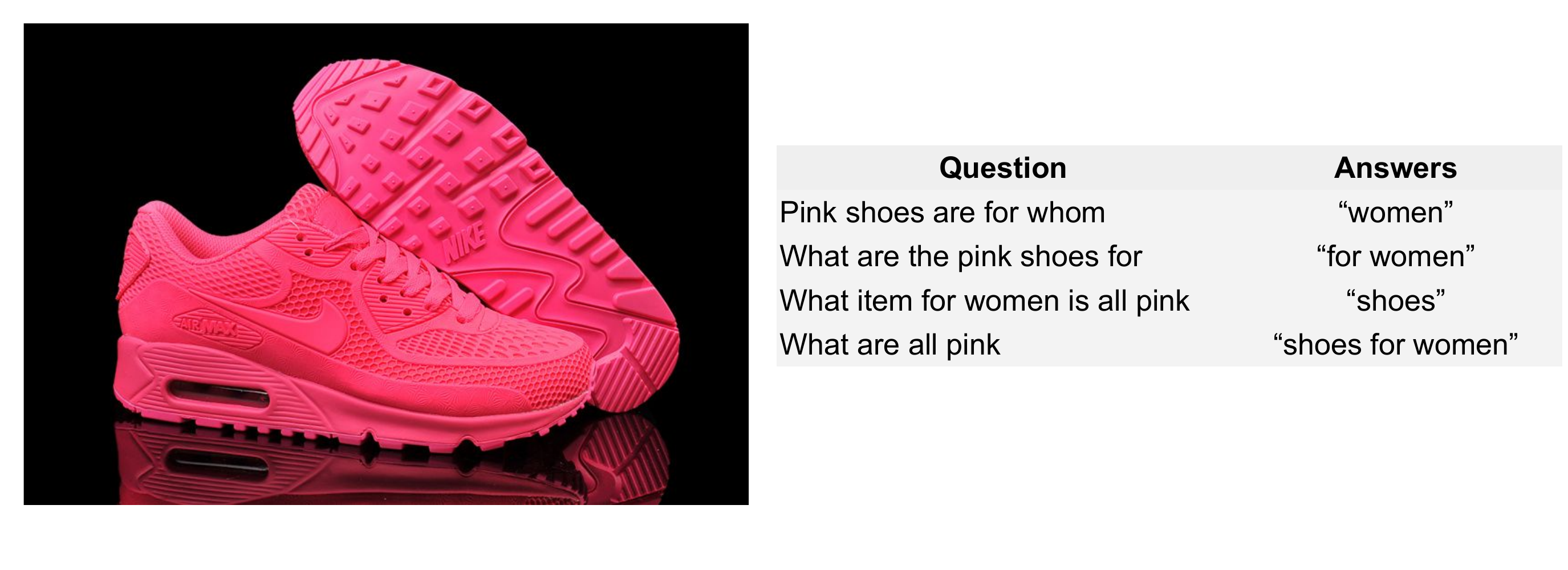}
}
\caption{\small Additional examples from \qsq{} COCO (top) and \qsq{} CC3M (bottom). Questions with the green background are present in \vqaset{}.}
\label{fig:vq2a_qual}
\end{figure*}

\section{Additional Examples and Analysis of Generated Data}
\label{apdx:data_analysis}

Fig.~\ref{fig:vq2a_qual} provides additional examples of \qsq{} COCO and CC3M generated VQA triplets, showing the diversity compared to what can be found in \vqaset.

\begin{table*}[t]
\centering
\small		
\begin{tabular}{|l|c|c|c|l|}
\hline
Question Prefix &  \vqaset \% & \qsqcoco \% & \qsqcc \% & Question Example from \qsqcoco\\
\hline
        \qex{What is} &       0.140 &             0.288 &             0.217 &                       \qex{What is the man swinging?} \\
       \qex{How many} &       0.110 &             0.022 &             0.005 &  \qex{How many people are standing in front of a tv?} \\
         \qex{Is the} &       0.098 &             0.084 &             0.053 &                \qex{Is the baby wearing a Santa hat?} \\
     \qex{What color} &       0.090 &             0.022 &             0.018 &                   \qex{What color is the man's hair?} \\
        \qex{Is this} &       0.082 &             0.008 &             0.015 &                      \qex{Is this a safe way to fly?} \\
       \qex{Is there} &       0.037 &             0.011 &             0.022 &                \qex{Is there a pool in the backyard?} \\
      \qex{What kind} &       0.025 &             0.049 &             0.078 &           \qex{What kind of truck is the yellow one?} \\
       \qex{What are} &       0.024 &             0.049 &             0.022 &   \qex{What are the sheep and other animals roaming?} \\
        \qex{Are the} &       0.024 &             0.022 &             0.007 &      \qex{Are the apples on the cutting board green?} \\
      \qex{Are there} &       0.020 &             0.002 &             0.004 &          \qex{Are there any exceptions to this rule?} \\
       \qex{Where is} &       0.019 &             0.071 &             0.034 &            \qex{Where is the tennis player pictured?} \\
      \qex{What type} &       0.018 &             0.006 &             0.022 &                   \qex{What type of picture is this?} \\
          \qex{Is it} &       0.017 &             0.001 &             0.005 &            \qex{Is it possible to eat a whole pizza?} \\
       \qex{Does the} &       0.014 &             0.007 &             0.007 &          \qex{Does the adult giraffe have any young?} \\
      \qex{What does} &       0.011 &             0.015 &             0.038 &      \qex{What does a giraffe do with its long neck?} \\
      \qex{Where are} &       0.006 &             0.032 &             0.014 &       \qex{Where are the skateboarders in the photo?} \\
         \qex{Who is} &       0.005 &             0.054 &             0.020 &                            \qex{Who is in the photo?} \\
        \qex{What do} &       0.002 &             0.003 &             0.018 &                \qex{What do the father and son ride?} \\
       \qex{What was} &       0.000 &             0.009 &             0.023 &                  \qex{What was the woman looking at?} \\
       \qex{What did} &       0.000 &             0.001 &             0.021 &                 \qex{What did the cat lay inside of?} \\
\hline
\end{tabular}
\caption{Most popular question prefix distribution on valid questions whose answers are in the 6k target vocabulary.}
\label{tab:coco_prefix_stats}
\end{table*}

Table~\ref{tab:coco_prefix_stats} presents the top question prefixes and their distribution in the \vqaset and \qsq-based dev sets, showing significant differences between datasets. Many questions in \vqaset are of boolean answer type, e.g. \qex{is the}, \qex{is there} and \qex{does the}, summing to 29.2\%. In addition, (\qex{how many}) questions are frequent, 11\%. Finally, questions for the color attribute are standing out with 9\%. 
On the other hand, \coco and \ccslong questions are more explanatory in nature, with the majority of questions (45.5\% in \coco, 43.9\% in \ccslong) of the form \qex{what is/are/do/does/type}. Another type that is more prominent in \coco and \ccslong are \qex{where is/are} questions, which are more than twice frequent compared to \vqaset.

\begin{figure}[t]
\centering
\small
\resizebox{\linewidth}{!}{%
\includegraphics{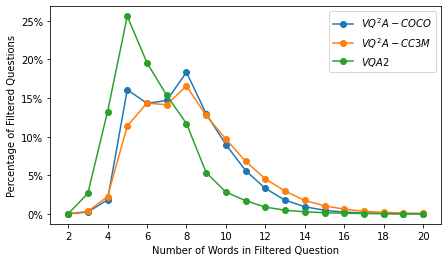}
}
\caption{Question length distributions per dataset.}
\label{fig:question_length_distribution}
\end{figure}

\begin{figure}[t]
\resizebox{\linewidth}{!}{%
\includegraphics{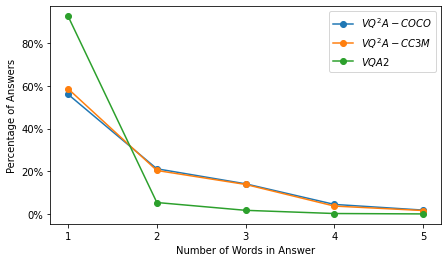}
}
\caption{Answer length distributions per dataset.}
\label{fig:answer_length_distribution}
\end{figure}

Another difference between the manually curated \vqaset dataset and the \qsq automatically generated datasets is question and answer word length distribution (Fig. \ref{fig:question_length_distribution} and \ref{fig:answer_length_distribution}). The questions in \qsqcc and \qsqcoco have an average word length of 8.3 and 7.8 respectively, while the average \vqaset is 6.3. Inspecting the generated questions, we noticed that QG model tends to quote parts of the caption,  extending the question length.
The average answer word length in \qsqcc and \qsqcoco is 1.76 and 1.85 words respectively, while in \vqaset it is 1.1. While all answers tend to be short, the \qsq-induced datasets have more ``detailed'' answers of length 2-3 words.  
\begin{figure}
\centering
\resizebox{\linewidth}{!}{%
\includegraphics{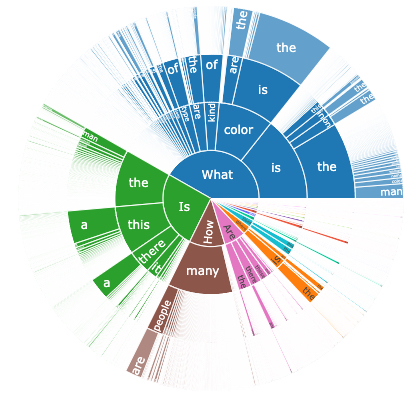}
}
\resizebox{\linewidth}{!}{%
\includegraphics{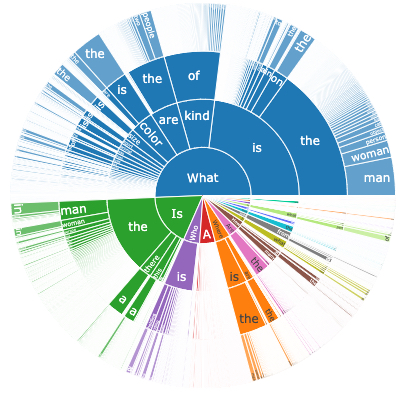}
}
\resizebox{\linewidth}{!}{%
\includegraphics{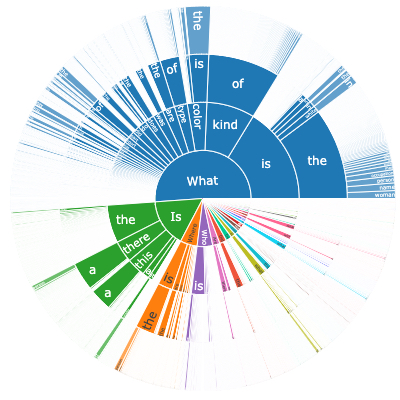}
}
\caption{\vqaset (top), \qsqcoco (middle), \qsqcc (bottom) sunburst plots of question prefixes.}
\label{fig:sunburst}
\end{figure}

Fig.~\ref{fig:sunburst} offers a more visual view of the differences between question type distribution presented in Table~\ref{tab:coco_prefix_stats}.

Table~\ref{tab:filtering_prefix_stats} depicts the percentage of questions of each type (prefix) that were retained (not filtered out) when applying the question answer validation phase of \qsq (Section~\ref{ssec:qa_filtering}).

\begin{table}[t]
\centering
\small		
\begin{tabular}{|l|c|c|}
\hline
\multicolumn{1}{|c|}{Question}          & \qsqcoco         &       \qsqcc               \\
\multicolumn{1}{|c|}{Prefix}            & Filter Pass Ratio  &       Filter Pass Ratio       \\ 
\hline
        \qex{What is} &                    0.73 &                    0.65 \\
         \qex{Is the} &                    0.64 &                    0.39 \\
      \qex{What kind} &                    0.84 &                    0.80 \\
       \qex{How many} &                    0.83 &                    0.51 \\
     \qex{What color} &                    0.92 &                    0.90 \\
       \qex{Where is} &                    0.79 &                    0.79 \\
        \qex{Is this} &                    0.83 &                    0.62 \\
       \qex{What are} &                    0.75 &                    0.71 \\
         \qex{Who is} &                    0.85 &                    0.79 \\
       \qex{Is there} &                    0.73 &                    0.47 \\
      \qex{What does} &                    0.75 &                    0.67 \\
        \qex{Are the} &                    0.58 &                    0.32 \\
      \qex{Where are} &                    0.80 &                    0.81 \\
      \qex{What type} &                    0.84 &                    0.81 \\
       \qex{What was} &                    0.72 &                    0.67 \\
       \qex{Does the} &                    0.60 &                    0.43 \\
      \qex{Are there} &                    0.80 &                    0.62 \\
        \qex{What do} &                    0.76 &                    0.72 \\
       \qex{What did} &                    0.69 &                    0.64 \\
          \qex{Is it} &                    0.62 &                    0.59 \\
\hline
\end{tabular}
\caption{Question filtering stats.}
\label{tab:filtering_prefix_stats}
\end{table}

\section{Implementation Details}
\label{apdx:implement}

\subsection{Details on Data Processing}
\label{apdx:implement_data_process}

Our default question and answer preprocessor is based on \cite{jiang2018pythia,singh2020mmf}\footnote{\url{https://github.com/facebookresearch/mmf/blob/main/mmf/datasets/processors/processors.py}}, with the exception of \gqa{} which we use \footnote{\url{https://github.com/stanfordnlp/mac-network/blob/gqa/preprocess.py}}. The unified answer vocabulary used in our experiments is the union of top answers from existing COCO-based VQA benchmarks: \vqaset{} (3,128, minimum answer frequency=9), \gqa{} (1,843, all), \okvqa{} (2,000, top), and \vsw{} (3,140, minimum answer frequency=3) of total size 5,971

For each image-unique question pair generated by our \qsq{} approach, we reduce or expand a list of possibly different candidate answers based on the list length, such that we eventually have a target list of answers of size 10. In particular, we first sort the answers based on their lengths ("dog" before "black dog"), and select up to top-10 answers. If the list legnth is less than 10, we replicate each of the top answers one-by-one until we have the list of size 10, similar to the process in \okvqa{}\cite{okvqa}. This is to ensure that we can adopt \emph{VQA Accuracy} to make the performance comparison.

\subsection{Details on Training and Evaluating Visual Question Answering}
\label{apdx:implement_vqa}

Our code for the VQA model is based on the Flaxformer framework\footnote{\url{https://github.com/google/flaxformer}}.
Both the text encoder and the multi-modal encoder have 6 blocks of Transformers, each of which consists of self-attention and a feed-forward network. We use 12 heads of inner dimension of 64, the embedding dimension of 768, and the MLP dimension of 2048.
During training, we use Adafactor~\cite{adafactor}, with an initial learning rate of 0.0025, a linear warm-up step of 5K for (pre-)training and 1K for fine-tuning, and an “inverse square root” learning rate schedule $\frac{1}{\sqrt{\max(n,k)}}$, where $n$ is the current training iteration and $k$ is the number of warm-up steps. We use a dropout rate of 0.0. We train each of the models with data parallelism using 16 Cloud TPU Pods\footnote{\url{https://cloud.google.com/tpu}}, each with a batch size of 256, unless otherwise stated.

The default numbers of training steps during training and fine-tuning are 100K and 30K, respectively. The exceptions are \okvqa{} (30K/15K) and \qsq{} CC3M (150K/NA). In addition, in the two-stage training where we fine-tune a \qsq{}-CC3M model with \qsq{} COCO, we also use 100K steps. Each single training run on average took fewer than 10 hours, including the time used to evaluate a checkpoint --- every 1K iterations. For instance, training on \vqaset{} took approximately 7 hours, \qsq{} COCO 13 hours, \qsq{} CC3M 10 hours. Note that \qsq{} COCO has larger evaluation set than other datasets, hence taking longer time to to train then \qsq{} CC3M. 

The hyperparameters for Transformers are selected to be consistent with a T5-base checkpoint, which has 220 million parameters \cite{t5} (except that now we have 2 encoders rather than an encoder and a decoder). We initially tuned the initial learning rate (0.0125, 0.075, 0025, 0.00125, 0.00075) and the dropout rate (0.0, 0.1, 0.2) on a fully-supervised model on \vqaset{} baseline using \emph{VQA Accuracy} and observed that 0.0025 and 0.0 work robustly across our experiments but we did not extensively tuned them in all of our experiments.

We implement \emph{VQA Accuracy} ourselves based on the official challenge page for \vqaset{}\footnote{\url{https://visualqa.org/evaluation.html}}.

\begin{table}[t]
\centering
\small		
{\setlength{\tabcolsep}{0.15cm}
\begin{tabular}{|l|c|c|c|}
\hline
\multicolumn{1}{|c|}{Question} & \vqaset  & \qsqcoco &  \qsqcc    \\
\multicolumn{1}{|c|}{Prefix}   & Supervised & Zero-shot &  Zero-shot  \\ 
\hline
\qex{is there}        &          98.6 &          98.1 &          98.2 \\
\qex{are there}       &          98.0 &          97.1 &          97.2 \\
\qex{does this}       &          98.0 &          95.1 &          95.8 \\
\qex{are they}        &          96.9 &          95.0 &          95.3 \\
\qex{does the}        &          96.4 &          95.2 &          95.9 \\
\qex{is it}           &          96.3 &          91.4 &          92.7 \\
\qex{is this}         &          96.1 &          91.2 &          92.8 \\
\qex{are the}         &          95.6 &          92.1 &          93.1 \\
\qex{is the}          &          95.3 &          91.7 &          92.9 \\
\qex{are these}       &          95.1 &          90.7 &          92.2 \\
\hline
\qex{what color}      &          69.2 &          64.8 &          56.8 \\
\qex{what kind}       &          56.3 &          35.8 &          31.4 \\
\qex{what type}       &          54.4 &          32.3 &          30.8 \\
\qex{what are}        &          51.3 &          40.2 &          33.9 \\
\qex{how many}        &          49.3 &          29.4 &          19.5 \\
\qex{what is}         &          48.5 &          39.4 &          32.2 \\
\qex{where are}       &          40.9 &          33.9 &          27.6 \\
\qex{where is}        &          35.1 &          26.0 &          23.0 \\
\qex{what does}       &          33.0 &          24.1 &          20.3 \\
\qex{what time}       &          23.6 &          11.9 &          12.7 \\
\hline
\end{tabular}}
\caption{Average accuracy (\%) on \vqaset for the most common question prefixes.}
\label{tab:detailed_prefix_accuracies}
\vspace{-15pt}
\end{table}

\section{Additional Results}
\label{apdx:add_results}

Table~\ref{tab:detailed_prefix_accuracies} offers the Accuracy of the supervised \vqaset model, as well as of the zero-shot \qsq models, on the \vqaset devset, split by most common question prefixes. The Table is sorted by the supervised model's Accuracy. It shows a several performance differences, first between all types of boolean questions, which all have high precision on all models, vs. other types, which show not only lower performance for all models, but also more significant performance drop between the supervised and zreo-shot models.

Table~\ref{tab:eval_vqa_all} shows the zero-shot performance of models when using all of the \qsq dev sets, not only the manually validated sample, for which Table~\ref{tab:eval_vqa} reports results. What we see is that the difference in performance on the whole \qsq dev sets (Table~\ref{tab:eval_vqa_all}) is similar in magnitude to that of the manually validated dev samples (Table~\ref{tab:eval_vqa}), and most importantly, it keeps the order of models in terms of capabilities/performance. We therefore suggests that the utility of the \qsq{} approach could go beyond training; it can be used as an automatic test-bed for VQA robustness, if not for absolute figures, for ranking models for robustness zero-shot capabilities. 

\begin{table}[t]
\small
\begin{center}
\begin{tabular}{@{}l|c|c|cc@{}}
 & \multicolumn{4}{c}{Evaluation Benchmark} \\  \cline{2-5}
\multicolumn{1}{c|}{Training} & COCO- & \vqaset{} & \qsq{}& \qsq{} \\ 
\multicolumn{1}{c|}{data} & QA & & COCO & CC3M \\ \hline
\cocoqa{} & {\color{gray} 70.3} & 11.7 & 11.5 & 3.7 \\ \hline
\vqaset & 35.9 & {\color{gray} 68.8} & 41.1 & 33.3 \\ \hline
\qsq{} COCO & \textbf{55.9} & \textbf{60.0} & {\color{gray} 71.2} & \textbf{49.3} \\
\qsq{} CC3M & 42.1 & 56.5 & \textbf{60.3} & {\color{gray} 69.5} \\ \hline
\end{tabular}
\end{center}
\vspace{-6pt}
\caption{\textbf{\qsq{} as evaluation data} for measuring robustness: VQA Accuracy when training on "row" and tested on "column"; diagonal (gray) numbers denote the supervised setting, non-diagonal numbers denote zero-shot cross-dataset setting. Best zero-shot is \textbf{bold}.}
\vspace{-8pt}
\label{tab:eval_vqa_all}
\end{table}


Table~\ref{tab:train_vqa_answer_subset} shows the effect of candidate answer types on the \vqaset{} performance. 
We train our model on \qsq{} COCO or \qsq{} CC3M subsets with questions with (i) noun answers, (ii) yes/no answers, (iii) answers containing color-related tokens based on a list of common colors from Wikipedia, and (iv) answers containing digits from 0 to 100. 
We then evaluate models trained on these subsets on \vqaset{} using \emph{VQA Accuracy} and the normalized version (by the percentage of evaluation questions with corresponding answer types. This highlights the importance of incorporating diverse answer candidates in our datasets.
We also observe that \qsq{} CC3M is on par with \qsq{} COCO on yes/no-answer questions but are behind on nouns, color, and number, which we attribute to their lower degree image-text relevance, less mentioning of colors (due to the style of alt-texts vs. captions), and digit substitution.

Table~\ref{tab:train_vqa_random_subset} shows the effect of scale on the \vqaset{} performance. We randomly sampled 10\%, 20\%, and 50\% of \qsq{} COCO or \qsq{} CC3M training data. We observe that the bigger the data, the higher the accuracy. However, the gain is diminishing. We identify improving the data generation process to achieve higher degree of diversity in the output as interesting future work.

\begin{table}[t]
\small
\begin{center}
\begin{tabular}{@{}l|c|c@{}}
\multicolumn{1}{c|}{Training data} & \multicolumn{2}{c}{\emph{VQA Accuracy} on \vqaset{}} \\ \cline{2-3}
& Standard & Normalized \\ \hline
\qsq{} COCO & 60.0 & 60.0 \\ \hline
\qsq{} COCO nouns & 10.5 & 32.5\\
\qsq{} COCO yes/no & 38.4 & 94.3\\
\qsq{} COCO color & 6.7 & 55.6\\
\qsq{} COCO number & 3.9 & 25.4\\ \hline \hline
\qsq{} CC3M & 56.5 & 56.5 \\ \hline
\qsq{} CC3M nouns & 8.8 & 27.2\\ 
\qsq{} CC3M yes/no & 38.4 & 94.4 \\ 
\qsq{} CC3M color & 6.0 & 49.5 \\ 
\qsq{} CC3M number & 3.4 & 22.1\\ \hline
\hline
\end{tabular}
\end{center}
\vspace{-6pt}
\caption{\textbf{Effect of candidate answer types} on the \vqaset{} performance.}
\label{tab:train_vqa_answer_subset}
\end{table}

\begin{table}[t]
\small
\begin{center}
\begin{tabular}{@{}l|c|c@{}}
\multicolumn{1}{c|}{Training data} & \multicolumn{1}{c}{\emph{VQA Accuracy}} \\
 & \multicolumn{1}{c}{on \vqaset{}} \\ \hline
\qsq{} COCO (100\%) & 60.0 \\ \hline
\qsq{} COCO (50\%) & 58.5 \\
\qsq{} COCO (20\%) & 56.7 \\
\qsq{} COCO (10\%) & 55.4 \\ \hline \hline
\qsq{} CC3M (100\%) & 56.5 \\ \hline
\qsq{} CC3M (50\%) & 55.8 \\ 
\qsq{} CC3M (20\%) & 54.8 \\ 
\qsq{} CC3M (10\%) & 53.8 \\  \hline
\hline
\end{tabular}
\end{center}
\vspace{-6pt}
\caption{\textbf{Effect of dataset sizes} on the \vqaset{} performance.}
\label{tab:train_vqa_random_subset}
\end{table}

\begin{table}[t]
\small
\begin{center}
\begin{tabular}{@{}l|c|c|c@{}}
 & \multicolumn{3}{c}{Evaluation Benchmark} \\  \cline{2-4}
\multicolumn{1}{c|}{Approach} & \vqaset{} & \gqa{} & \okvqa{} \\ \hline \hline
\multicolumn{4}{c}{Zero-shot} \\ \hline
questions \qsq{} COCO & 48.9 & 44.4 & 11.4 \\
questions \qsq{} CC3M & 47.8 & 44.6 & 11.9 \\ \hline
\qsq{} COCO & 60.0 & 51.3 & 18.0 \\
\qsq{} CC3M & 56.5 & 49.9 & 19.1\\
\hline
\end{tabular}
\end{center}
\caption{\textbf{Zero-shot question-only baselines} using \qsq{} as training data.}
\vspace{-10pt}
\label{tab:train_vqa_qonly}
\end{table}

Table~\ref{tab:train_vqa_qonly} provides question-only baselines (no image features as input). Interestingly, the models trained on our generated \qsq{} data has similar answer distributions to those of existing VQA benchmarks. At the same time, this reveals the exploitation of the language bias, suggesting that additional research on bias mitigation is needed, both in terms of model and data (existing benchmarks and our datasets).

\section{Further Considerations}
\label{apdx:consider}

\mypartop{Information that names or uniquely identifies individual people or offensive content} COCO Captions are human-curated and cleaned while the approach to collection of CC3M upholds rigorous privacy and ethics standards such as the removal of offensive content and hypernymization. This significantly mitigates the risks that our \qsq{} datasets would contain such information.

\mypar{Intended uses} Due to considerations and limitations as we mention in Section~\ref{sec:discuss}, COCO Captions, CC3M, and our induced \qsq{} are intended to be used for research-only purposes.

\end{document}